\documentclass[11pt,a4paper]{article}
\usepackage[hyperref]{acl2021}
\usepackage{times}
\usepackage{latexsym}
\usepackage{amsfonts}
\usepackage{dsfont}
\usepackage{amssymb}
\usepackage{amsmath}
\usepackage{pgfplots}
\pgfplotsset{compat=1.16}
\usepackage{booktabs}
\usepackage{diagbox}
\usepackage{graphicx}
\usepackage{subfigure}
\usepackage{xspace}

\usepackage[textsize=scriptsize]{todonotes}

\usepackage{microtype}

\aclfinalcopy 
\def\aclpaperid{1511} 

\newcommand{\be}{\texttt{Bird's Eye}\xspace}
\newcommand{\we}{\texttt{Worm's Eye}\xspace}

\title{\textit{Bird's Eye}: Probing for Linguistic Graph Structures\\ with a Simple Information-Theoretic Approach}

\author{Yifan Hou \\
Department of Computer Science\\
  ETH Z\"{u}rich \\
  \texttt{yifan.hou@inf.ethz.ch} \\\And
  Mrinmaya Sachan \\
  Department of Computer Science\\
  ETH Z\"{u}rich \\
  \texttt{mrinmaya.sachan@inf.ethz.ch} \\}

\date{}

\begin{document}
\maketitle
\begin{abstract}
NLP has a rich history of representing our prior understanding of language in the form of graphs.
Recent work on analyzing contextualized text representations has focused on hand-designed probe models to understand how and to what extent do these representations encode a particular linguistic phenomenon.
However, due to the inter-dependence of various phenomena and randomness of training probe models, detecting how these representations encode the rich information in these linguistic graphs remains a challenging problem.
In this paper, we propose a new information-theoretic probe, \be, which is a fairly simple probe method for detecting if and how these representations encode the information in these linguistic graphs.
Instead of using classifier performance, our probe takes an information-theoretic view of probing and estimates the mutual information between the linguistic graph embedded in a continuous space and the contextualized word representations.
Furthermore, we also propose an approach to use our probe to investigate localized linguistic information in the linguistic graphs using perturbation analysis. We call this probing setup \we.
Using these probes, we analyze BERT models on their ability to encode a syntactic and a semantic graph structure, and find that these models encode to some degree both syntactic as well as semantic information; albeit syntactic information to a greater extent. Our implementation is available in \url{https://github.com/yifan-h/Graph_Probe-Birds_Eye}.
\end{abstract}



\section{Introduction}

Graphs have served as a predominant representation for various linguistic phenomena in natural language~\citep{dataset_ptb,Marneffe:2006,hockenmaier2007ccgbank,hajic-etal-2012-announcing,abend2013universal,AMR,bos2017groningen}. These graph based representations have served our intuition for representing both language structure~\citep{chomsky1957logical} as well as meaning~\citep{koller2019graph}.

With the growing popularity of pretrained language models that build contextualized text representations~\cite[inter alia]{elmo, bert}, various probing models have been introduced to understand if and how our linguistic intuitions are encoded in these representations. These probes train supervised models to predict pieces of linguistic information such as POS (part-of-speech), morphology, syntactic and semantic relations, and other local or long-range phenomena in language~\citep{pos_probe,morphology_probe, structural_probe, edge_probe, bert_probe}. 
However, it is still an open question if these representations somehow encode entire linguistic graph structures such as dependency and constituency parse trees or graph structured meaning representations such as AMR (Abstract Meaning Representation), UCCA (Universal Conceptual Cognitive Annotation), etc.

A popular recent work, the \textit{structural probe}~\citep{structural_probe}, has investigated how contextualized representations encode syntax trees.
%
They tested if a linear transformation of the network's word representation space can predict \textit{particular features} of the syntax tree, namely, the \textit{distance between words} and \textit{depth of words} in the tree. Thus, the structural probe cannot by itself answer the question if these representations encode entire linguistic graph structures. Moreover, the structural probe is only designed for tree structures and cannot be extended to general graphs.

\begin{figure*}[ht]
	\centering
	\includegraphics[scale=0.47]{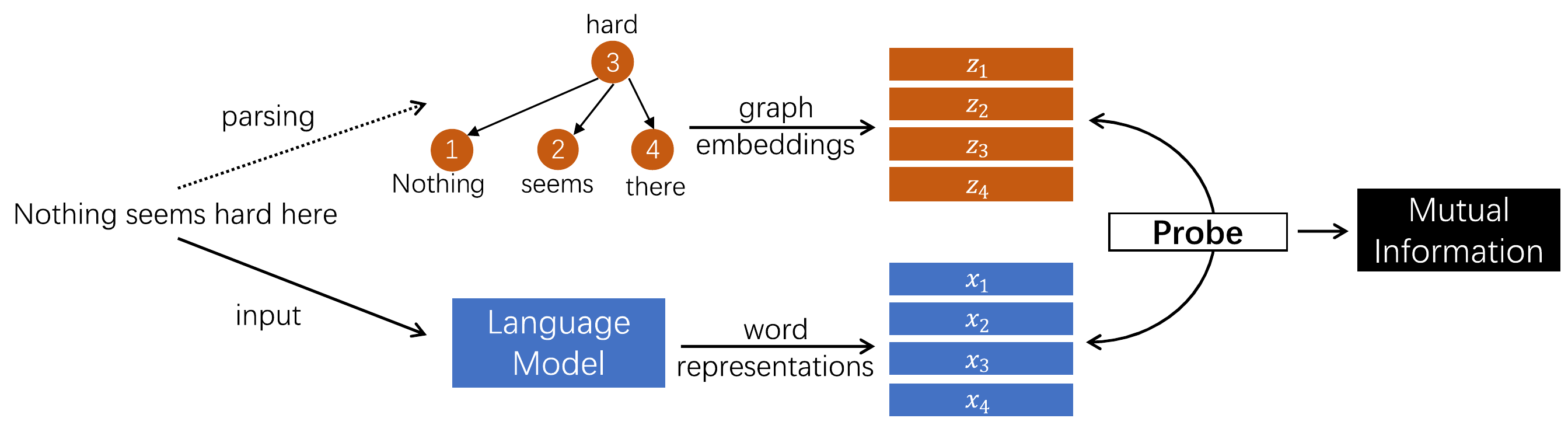}
	\vspace{-3.mm}
	\caption{Methodology of \be: To probe pretrained language models, linguistic graphs are embedded in a continuous space and the mutual information between graph embeddings and word representations is calculated.}
	\label{fig:probe_method}
	\vspace{-5.mm}
\end{figure*}

In this work, we introduce a new probing approach, \be, which can be used to detect if contextualized text representations encode entire linguistic graphs. \be is a simple information-theoretic probe~\citep{mi_probe} which first encodes the linguistic graph into a continuous representation using graph embedding approaches~\citep{hongyun_ge_survey} and then, estimates the mutual information between the linguistic graph representation space and the contextualized word representation space.
An illustration of the probe approach is given in Figure \ref{fig:probe_method}.
The information theoretic approach is more reliable than training a probe and using accuracy for probing as it is debatable if the classifier-based probe is probing or trying to solve the task~\citep{hewitt_control_task,mi_probe}.
We further extend \be to probe for localized linguistic information in the linguistic graphs such as POS or dependency arc labels in dependency parses. We call this probe, \we.

In our experiments, we first illustrate the reliability of our probe methods and show the randomness of previous probe methods that use accuracy. Then, we use \be to detect syntactic and semantic structures in BERT, showing that much syntactic and some semantic structure are encoded in BERT. Besides, we also use \we to probe for specific linguistic information in syntactic trees and semantic graphs respectively to see which kinds of localized linguistic information is encoded in BERT. Our probing results are consistent with previous probe methods~\citep{structural_probe, emily_geometric_bert,bert_no_semantic, tenney-etal-2019-bert, edge_probe, wu2020infusing}.
We also discuss limitations of our probe and how future work can build upon our foundation.

\section{\be Probe}
In this section, we introduce our information-theoretic approach for probing linguistic graph structures in word representations.
The MI estimate is used to understand how much of the information in the linguistic graph structure has been learnt by the pretrained models.


Let $X = \{x_1, \dots , x_T\}$ denote an input sentence (each $x_i$ is the contextual embedding of a token in the given vocabulary $\mathcal{V}$) and $G$ denote the corresponding linguistic graph.
Furthermore, let $\mathcal{X}$ denote a random variable that takes values ranging over all possible token sequences in $\mathcal{V}$. Correspondingly, let $\mathcal{G}$ denote a random variable that ranges over all possible corresponding linguistic graphs.
We use ${ I}(\mathcal{X};\mathcal{G})$ to denote the linguistic structure information that is included in the given word representations. Note that the MI value $I(\mathcal{X};\mathcal{G})$ is always non-negative, and a large MI implies that more of the structure information is encoded in the word representations.
In order to make the MI computation easier, we additionally assume alignments between the nodes $V$ in the graph $G$ and the words in $X$. This alignment is one to one, for example, in dependency parsing~\citep{dataset_ptb} but an aligner might be needed in some cases~\citep{AMR}.

There are three main challenges in estimating MI in our setting. 
First, the MI estimation of discrete graphs and continuous features has been an elusive problem~\citep{mi_estimation_discrete_continuous, kraskov2004_mi_estimation_nn, francisco_mi_graphs}, since there is no widely accepted definition of mutual information in this setting.
Second, the dimensionality of the contextualized word representations is very high. Traditional methods~\citep{moon1995_mi_estimation_kernel, steuer_mi_estimation_histogram, paninski_mi_estimation_histogram} for MI estimation do not scale well with large sample size or dimension~\citep{gao_mi_scale}. Getting accurate estimates of mutual information in the high dimension is not easy.
Third, graphs across different linguistic formalisms could have different entropy values, and thus the MI value ${ I}(\mathcal{X};\mathcal{G})$ may be uncomparable across the different linguistic graph formalisms. For example, if syntactic trees $\mathcal{G}$ and semantic graphs $\mathcal{G}'$ have the same MI value with $\mathcal{X}$ i.e. ${ I}(\mathcal{X};\mathcal{G})={ I}(\mathcal{X};\mathcal{G}')$ while the entropy values are fairly different i.e. ${ I}(\mathcal{X};\mathcal{G}) \approx { H}(\mathcal{G}) << { H}(\mathcal{G}')$, it is not proper to conclude that $\mathcal{X}$ contains the same amount of information from structures $\mathcal{G}$ and $\mathcal{G}'$, since they correspond to different percentages of the amount of uncertainty. Thus, the MI values must be interpreted carefully.

\be tackles the aforementioned difficulties by transforming the linguistic graphs into a continuous space using a graph embedding approach. Then the MI between graph embeddings and word representations is estimated using a recently proposed method~\citep{mine} which performs well even in high dimensions. Finally, we also estimate a lower and upper bound of the MI, which is used to interpret the MI value. We describe various stages of \be below:

\subsection{Graph Embedding}
The provided linguistic graphs can typically be represented as an adjacency matrix.
Directly calculating MI with the adjacency matrix is not useful due to the sparsity and discreteness of the adjacency matrix representation.
Thus, we transform the graphs into a continuous space where each node is represented by a continuous representation of same dimensionality.

Theoretically, if the graph embedding approach is perfect, we can use the invariant property of mutual information~\citep{kraskov2004_mi_estimation_nn}. 
This property states that under some fairly strong conditions, there exists an invertible function $f(\cdot)$ that satisfies $\mathcal{G}=f^{-1}(f(\mathcal{G}))$, where the graph embeddings are $\mathcal{Z}=f(\mathcal{G})$. Thus, we can transform $\mathcal{G}$ into graph embeddings $\mathcal{Z}$, and: 
\begin{equation}
\begin{aligned}
    { I}(\mathcal{X};\mathcal{Z}) \approx { I}(\mathcal{X};\mathcal{G})
\end{aligned}
\label{eq:invariance}
\end{equation}


In this paper, we use DeepWalk~\citep{deepwalk}, which is based on the skip-gram model~\citep{skipgram, word2vec} for graph embeddings\footnote{Note that the \be probe is a general probe. Other graph embedding approaches can also be selected for the transformation under specific conditions.}. Specifically, given a node $v \in V$ encoded as the one-hot vector $\mathds{1}_{v}$, the model tries to predict its neighbor's vector $\mathds{1}_{v'}$ where $v' \in N_{v}$. The graph $G=\{V, E\}$ is first sampled to generate a set of random walks. Then the graph neighborhood relationship is represented by the co-occurrence of nodes in the walk paths. Finally, for all the walks, Word2vec~\citep{word2vec} with skip-gram is used to maximize the co-occurrence likelihood\footnote{Note that in Word2vec, the window size is a hyperparameter that need to be selected by users. Here, for simplicity, we set the window size as $1$.}:
\begin{equation}
\begin{aligned}
    \mathcal{L}(\theta) = \prod_{v \in V} \prod_{v' \in N_v} \mathbb{P}(\mathds{1}_{v'}|\mathds{1}_{v};\theta).
\end{aligned}
\end{equation}
Let ${Z} = \oplus\{z_v | v \in V\}$ denote the learnt graph embedding where $z_v$ is the embedding of node $v$. Here $\oplus$ denotes the concatenation operation.

In our experiments, we also explore to what extent the original linguistic graphs can be restored by the graph embeddings, which tests the extent to which eq.~\ref{eq:invariance} holds and if we can use ${ I}(\mathcal{X};\mathcal{Z})$ instead of ${ I}(\mathcal{X};\mathcal{G})$ to estimate MI. More details can be found in Appendix~\ref{appendix:ge}

\subsection{Mutual Information Estimation}
To estimate ${ I}(\mathcal{X};\mathcal{Z})$ in high dimensions, we maximize the compression lemma lower bound~\citep{bound} as mentioned in~\newcite{mine}. Specifically, for a pair of random variables $\mathcal{X}$ and $\mathcal{Z}$, the mutual information is equivalent to the Kullback-Leibler (KL) divergence between the joint distribution $\mathbb{P}_{\mathcal{X}\mathcal{Z}}$ and the product of the marginal distributions $\mathbb{P}_{\mathcal{X}} \otimes \mathbb{P}_{\mathcal{Z}}$:
\begin{equation}
\begin{aligned}
    { I}(\mathcal{X};\mathcal{Z}) = { D_{KL}}(\mathbb{P}_{\mathcal{X}\mathcal{Z}}||\mathbb{P}_{\mathcal{X}} \otimes \mathbb{P}_{\mathcal{Z}}).
\end{aligned}
\end{equation}
From the compression lemma lower bound~\citep{bound}, the KL divergence $D_{KL}(\mathbb{P}||\mathbb{Q})$ can be bounded as:
\begin{equation}
\begin{aligned}
    D_{KL}(\mathbb{P}||\mathbb{Q}) \ge \sup_{T \in \mathcal{F}} \mathbb{E}_{\mathbb{P}}[T] - \log(\mathbb{E}_{\mathbb{Q}}[e^T]), 
\end{aligned}
\label{eq:kl_bound}
\end{equation}
where $\mathcal{F}$ can be any class of functions $T: \Omega \rightarrow \mathbb{R}$ satisfying certain integrability constraints. Thus, in the inequality~\ref{eq:kl_bound}, the lower bound can be obtained by finding a function in the set $\mathcal{F}$:
\begin{equation}
\begin{aligned}
    I(\mathcal{X};\mathcal{Z}) \ge \sup_{T \in \mathcal{F}} \mathbb{E}_{\mathbb{P}_{\mathcal{X}\mathcal{Z}}}[T] - \log(\mathbb{E}_{\mathbb{P}_{\mathcal{X}} \otimes \mathbb{P}_{\mathcal{Z}}}[e^T]).
\end{aligned}
\notag
\end{equation}
To get a tight estimate of $I(\mathcal{X};\mathcal{Z})$, we need the lower bound to be as high as possible. Thus, the MI estimation problem turns into an optimization problem to maximize the compression lemma lower bound. To ensure that, similar to ~\newcite{mine}, we let $\mathcal{F} = \{T_{\theta}\}_{\theta \in \Theta}$ be the set of functions parametrized by a neural network, and optimize the neural network using stochastic gradient descent. Formally, the objective function is:

\begin{equation}
\begin{aligned}
     \max_{\theta \in \Theta} ( \mathbb{E}_{\mathbb{P}^{(n)}_{\mathcal{X}\mathcal{Z}}}[T_{\theta}] - \log(\mathbb{E}_{\mathbb{P}^{(n)}_{\mathcal{X}} \otimes \mathbb{P}^{(n)}_{\mathcal{Z}}}[e^{T_{\theta}}])).
\end{aligned}
\label{eq:obj}
\end{equation}
Here, $\mathbb{P}_{\mathcal{X}\mathcal{Z}}^{(n)}$, $\mathbb{P}_{\mathcal{X}}^{(n)}$ and $\mathbb{P}_{\mathcal{Z}}^{(n)}$ are empirical joint and marginal distributions over a sample of $n$ (sentence, graph) pairs. 


We calculate graph embeddings for each sentence independently, and regard one sentence as a mini-batch to optimize the neural network iteratively for MI estimation. Note that different from existing probe models, our objective of the neural network is to find an optimal function in $\mathcal{F}$ and estimate MI, rather than use prediction accuracy. Besides, the neural network is very simple (MLP). Therefore, there is no need to split dataset into training and test to test generalization in MI estimation\footnote{Alternatively, the dataset can be divided evenly into training and test for MI estimation.}. The negative of the training loss as eq.~\ref{eq:obj} can be taken as MI estimation directly~\citep{mine, mi_estimation_loss}.
%
In our experiments, we verify the effectiveness of the MI estimation method to prove that the probe is stable. More technical details of the MI estimation model and how it is trained are given in Appendix~\ref{appendix:mi}.


\subsection{Control Bounds}
Next, we introduce two control bounds to interpret the MI value, whose functions are similar to the control task introduced by~\citet{hewitt_control_task}. As mentioned, comparing MI alone across different types of structures is not useful, since the entropy values of graph embeddings can also be different.
Thus, we calculate an upper and a lower bound of the MI value based on the graph structures. Instead of using the MI value alone, we interpret it by its relative value in terms of the two control bounds.
Formally, for the MI between graph embeddings and word representations, we have: 
\begin{equation}
\begin{aligned}
     I(\mathcal{R};\mathcal{Z}) \leq I(\mathcal{X};\mathcal{Z}) \leq I(\mathcal{Z};\mathcal{Z}).
\end{aligned}
\label{eq:two_bounds}
\end{equation}
The lower bound is the MI between a truly random variable $\mathcal{R}$ (i.e., independent of the graph) and the graph embedding $\mathcal{Z}$. Thus, $I(\mathcal{R};\mathcal{Z}) = 0$. The upper bound telescopes to the graph structure's self-entropy\footnote{Note that for continuous random variables $\mathcal{Z}$, the number of values that $\mathcal{Z}$ can take is infinite. In this condition, $I(\mathcal{Z};\mathcal{Z})$ tends to infinity. Thus, we use a small noise $\epsilon$ and approximate it as $I(\mathcal{Z}+\epsilon,\mathcal{Z})$.}.

Using these two control bounds, we interpret the structure information by the relative MI value\footnote{The definition is similar to the uncertainty coefficient.}:
\begin{equation}
\begin{aligned}
     MIG(\mathcal{G}) = \frac{\hat{I}(\mathcal{X};\mathcal{Z})-\hat{I}(\mathcal{R};\mathcal{Z})}{\hat{I}(\mathcal{Z};\mathcal{Z})-\hat{I}(\mathcal{R};\mathcal{Z})},
\end{aligned}
\label{eq:ratio}
\end{equation} 
The MI estimates $\hat{I}(\mathcal{Z};\mathcal{Z})$ and $\hat{I}(\mathcal{R};\mathcal{Z})$ can be obtained in the same way as $\hat{I}(\mathcal{X};\mathcal{Z})$ (using the MI estimation method mentioned above). MIG (eq.~\ref{eq:ratio}) scales the MI value for graph embeddings with different self-entropy values into the same range: $MIG(\mathcal{G}) \in [0,1]$. Intuitively, MIG captures what percentage of the structure information is encoded in word representations.

Since $MIG(\mathcal{G})$ is scaled using $I(\mathcal{R};\mathcal{Z})$, it also helps reduce the error in MI estimation. As mentioned, we maximize compression lemma lower bound~\ref{eq:obj} as the MI estimate. However, there could be a gap between it and the ground-truth MI value. Based on the fact that the ground-truth $I(\mathcal{R};\mathcal{Z})=0$, we can know that the gap $I(\mathcal{R};\mathcal{Z}) - \hat{I}(\mathcal{R};\mathcal{Z})$ is equal to $- \hat{I}(\mathcal{R};\mathcal{Z})$. In MIG (eq.~\ref{eq:ratio}), the gap is added for both numerator and denominator, which reduces the error brought by MI estimation\footnote{In our experiment, we show that the estimated values satisfy $|\hat{I}(\mathcal{R};\mathcal{Z})| < 10^{-3} \times \hat{I}(\mathcal{Z};\mathcal{Z})$, which is small enough to be ignored.}. 

\subsection{\we Probe for Localized Linguistic Structure}
\be allows us to probe for entire linguisitic structures. However, for us to have a complete understanding, we might also want to probe for some localized information in the linguistic graphs. For example, we may want to know if BERT knows about POS tags or certain dependency relations in the syntax parse.
We formulate probing for localized linguistic information as probing for a subgraph of the linguistic graph and reuse our \be probe for it. We call this setting \we as we are now analyzing if these representations capture local sub-structures.

To probe localized linguistic information $G_s=\{V_s, E_s\}$, we use perturbation of the original structure for analysis. 
Specifically, we add a perturbation to the original graph embedding ${Z}$ based on the subgraph $G_s$. For all the nodes in $V_s$ or nodes connected by edges in $E_s$, we add a noise on their corresponding node representations in ${Z}$. Let ${Z}'$ denote the corrupted graph embedding.
Then, we define the following:
\begin{equation}
\begin{aligned}
     MIL(\mathcal{G}_s) = 1- \frac{\hat{I}(\mathcal{X};\mathcal{Z}')-\hat{I}(\mathcal{R};\mathcal{Z})}{\hat{I}(\mathcal{X};\mathcal{Z})-\hat{I}(\mathcal{R};\mathcal{Z})},
\end{aligned}
\label{eq:ratio_local}
\end{equation} 
MIL describes how much MI is contributed by the local structure $\mathcal{G}_s$. When the local structure is the whole graph, $\mathcal{Z}'$ is completely noisy and $MIL(\mathcal{G}_s)$ equals to $1$, which means the entire MI value $I(\mathcal{X};\mathcal{Z})$ is contributed by the local structure. If the local structure is an empty set, we have $\mathcal{Z}'=\mathcal{Z}$. Then we can get $MIL(\mathcal{G}_s)=0$, representing that the local structure does not contribute anything to the MI value.

If we control the perturbation of different types of local structures at the same level, we can compare how well they are captured by the word representations relative to each other using eq.~\ref{eq:ratio_local}. Specifically, for relations with labels, e.g., types of dependency relations in syntax trees, we set the same perturbation on the graph embeddings. Then, we test and compare $MIL(\mathcal{G}_s)$ for different types of relations. Larger $MIL(\mathcal{G}_s)$ for a particular relation type implies that more information about this relation type is encoded in the word representations.

\section{Probing for Syntactic and Semantic Graph Structures}
We use our \be probe to detect two linguistic structures in the pretrained models, namely, dependency syntax~\citep{dataset_ptb,Marneffe:2006} and a more semantic formalism, AMR~\citep{AMR}.

We first use our model to probe for Stanford dependencies~\citep{SDF}. For a sentence $X$ with tokens $\{x_1, x_2,.. x_T\}$, the syntax tree defines a directed labelled tree where tokens $x_i$ are represented as nodes and relations among them as labeled edges. We ignore the edge direction and labels for simplicity in our work\footnote{The Stanford dependency tree also contains one empty root node, which is also ignored}. Future work can consider incorporating edge direction and labels.
We embed the given syntax tree into a continuous space as mentioned before.
Then, we calculate the
MIG (eq. \ref{eq:ratio}) as described before to determine how much syntax information is captured in the given contextualized representations.

Next, we test if contextualized representations capture a semantic graph representation -- the Abstract Meaning Representation (AMR)~\citep{AMR}. Different from syntactic trees, semantic graphs are not tree structured, and there can be loops or reentrencies. In the AMR annotation, plurality, articles and tenses were dropped and thus, there is no one-to-one corresponding between words in the sentence and nodes in the AMR graph. Thus, we use an off-the-shelf aligner~\citep{nima_amr_aligner} and calculate MI between the AMR graph embedding and the representations of those words that are aligned with a node in the AMR graph.
For simplicity, edge directions and labels are also ignored in this setting.


\section{Experiments}
Our experiments mainly comprise of two parts:

\noindent{\bf 1. Verification of the probe:} The first part is for verification of the probing methodology and ensuring that the graph embeddings retain information about the linguistic graphs i.e eq.~\ref{eq:invariance} holds. We do this by testing if the graph embeddings can be used to restore the original graph. 

\noindent{\bf 2. Probing for graph structures:} The second part is about using the probe to detect syntactic and semantic graph structures in BERT. Importantly, we probe if pretrained BERT captures entire graph structures as well as specific relational information in these linguistic graphs. 
To contrast with previous accuracy and training based probes, we also train a group of simple MLP models with different number of hidden layers and use accuracy for probing. We show that designing and training a model to probe entire or localized linguistic structures is not as reliable as our information-theoretic approach.



We use gold annotations from the Penn Treebank and the AMR Bank for all our experiments.
For the contextualized word representations, we select pretrained BERT models, specifically BERT-base (uncased) and BERT-large (uncased). Since BERT generates word-piece embeddings, to align them with gold word-level tokens, we represent each token as the average of its word-piece embeddings as in~\newcite{structural_probe}. We also use two non-contextual word embeddings as baselines: GloVe embeddings~\citep{jeffrey_glove} and ELMo-0, character-level word embeddings with no contextual information generated by pretrained ELMo~\citep{elmo}.

\subsection{Evaluation of Graph Embeddings}
We first evaluate how well the graph embeddings can capture the linguistic graph structures by predicting the original graphs with them. We use simple MLPs of $6$ different settings with varying number of hidden layers.
More details can be found in Appendix~\ref{appendix:mlp}. We use AUC score as the metric to evaluate the graph prediction performance, which is a common metric in link prediction that computes area under the ROC curve~\citep{tom_auc}. 

\begin{table}[thbp]
\center
	\caption{Performance (AUC score for link prediction) on restoring graphs with graph embeddings}
	\label{tab:ge_test}
	\vspace{-2mm}
	\scalebox{0.82}{
		\begin{tabular}{c|c|c}
			\hline
			\toprule
			\# of hidden layers & Syntax tree & AMR graph \\
			\midrule
			\hline
			0 & 0.5620 & 0.5620 \\
			1 & 0.6330 & 0.5494 \\
			2 & 0.9637 & 0.8804 \\
			3 & 0.9780 & 0.9263 \\
			4 & 0.9806 & 0.9162 \\
			5 & 0.9791 & 0.9192 \\
			\bottomrule
			\hline
		\end{tabular}}
		\vspace{-2.mm}
\end{table}

\begin{figure*}[!htbp]
\centering
\subfigure{
\begin{minipage}[t]{0.485\linewidth}
\centering
	\includegraphics[scale=0.25]{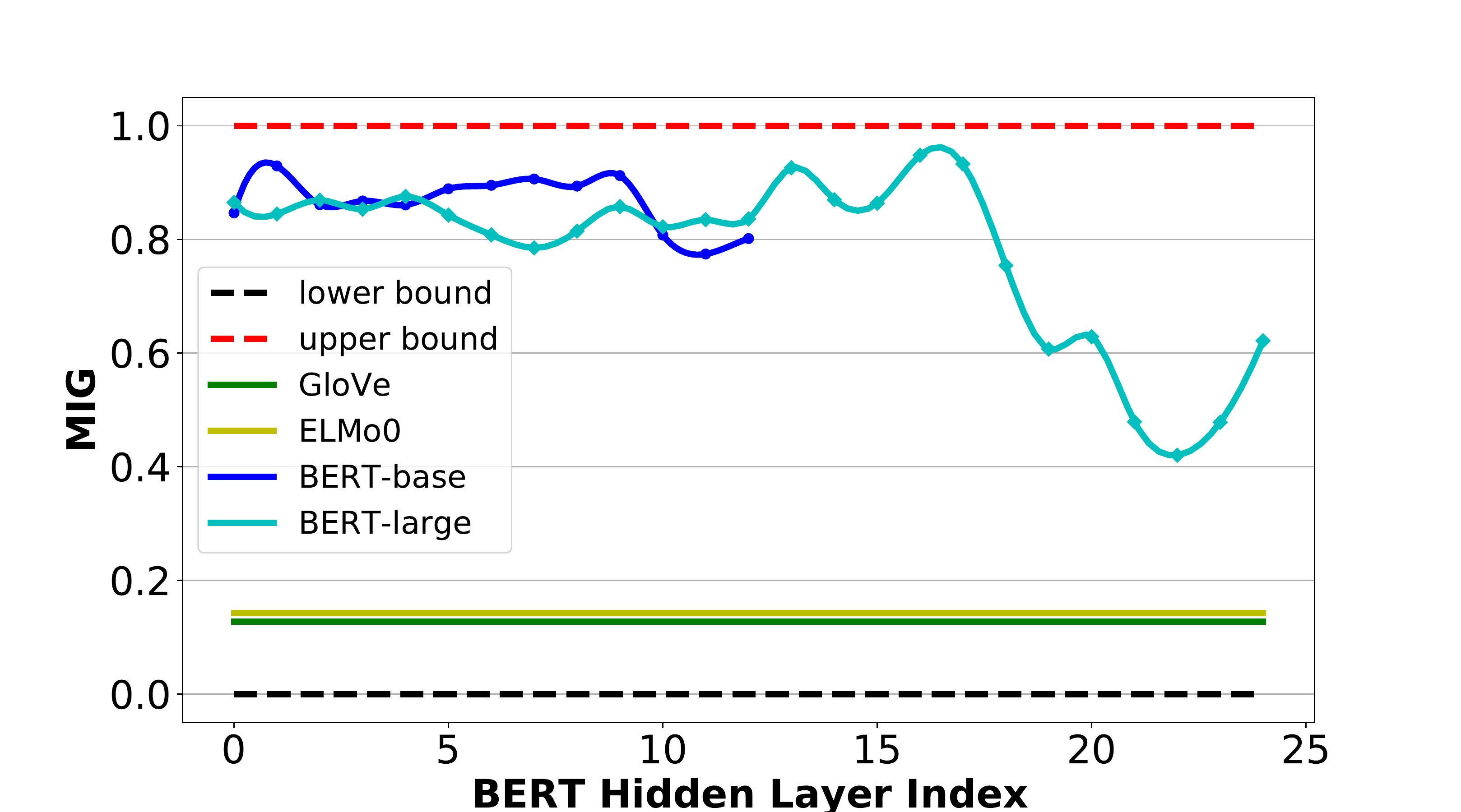}
	\vspace{-6.mm}
	\label{fig:mi_layers_ptb}
	\vspace{-2.mm}
\end{minipage}
}
\subfigure{
\begin{minipage}[t]{0.485\linewidth}
\centering
	\includegraphics[scale=0.25]{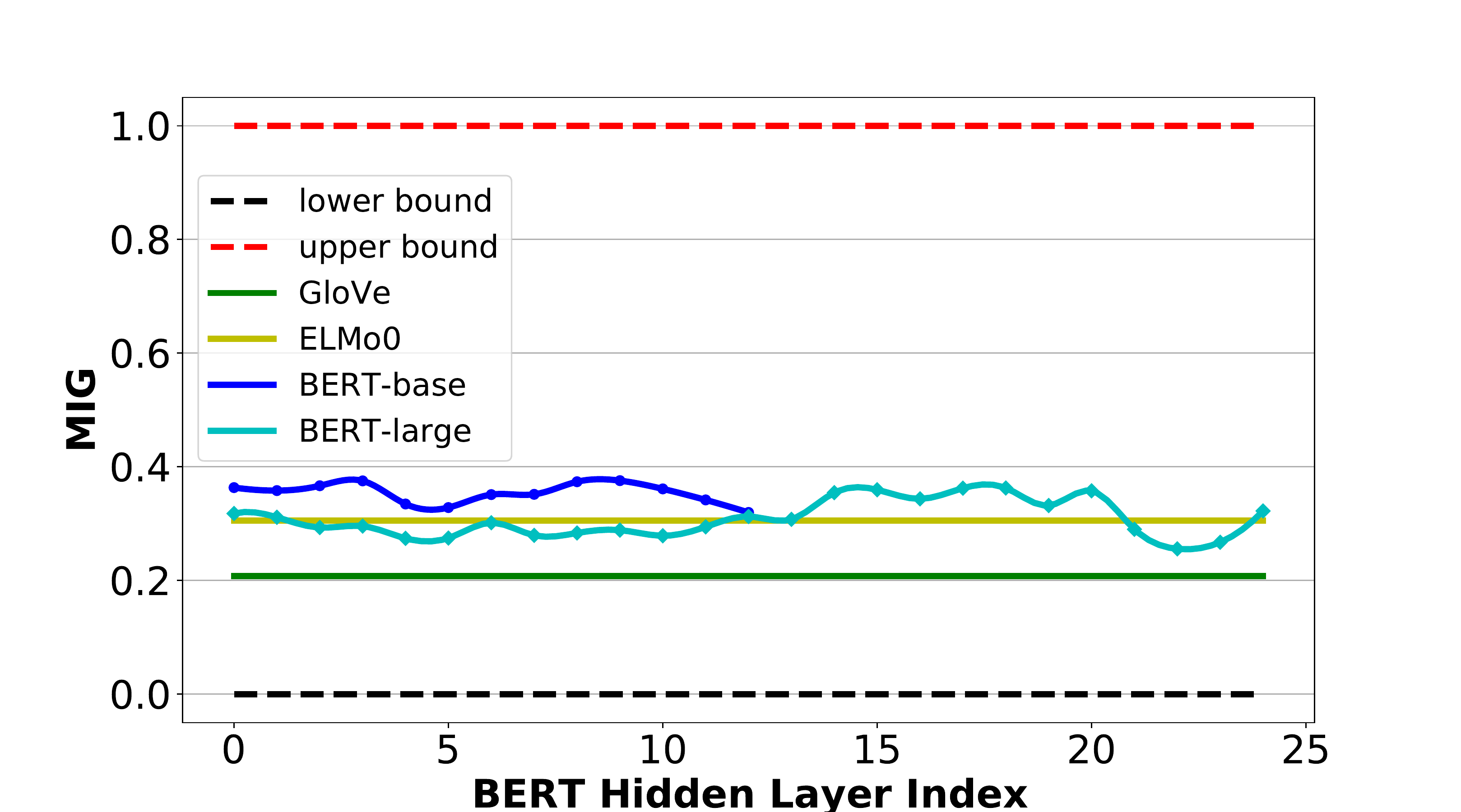}
	\vspace{-6.mm}
	\label{fig:mi_layers_amr}
	\vspace{-2.mm}
\end{minipage}
}
\centering
\caption{$MIG$ scores with syntactic and semantic structures, respectively for word representations in BERT models (BERT-base with 12 layers and BERT-large with 24 layers). Note that results at the input layer are also reported, where the BERT Hidden Layer Index is $0$).}
\end{figure*}

The results are presented in Table~\ref{tab:ge_test}. We can see that for both syntax trees and semantic graphs, MLPs can achieve good performance in restoring the original graph using graph embeddings where the AUC score is quite high. Thus, we can be confident that equation~\ref{eq:invariance} holds, and we can calculate MI based on the graph embeddings. Future work can explore better graph embedding approaches. 
We also evaluate our probe by adding noisy representations to the graph embeddings to prove that it is capable of teasing out different levels of dependencies. Details can be referred to in Appendix~\ref{appendix:mi_estimation}.

\subsection{Probing Entire Structures}
We first used the \be probe to detect if entire linguistic structures are encoded in hidden representations of BERT\footnote{For all MI estimation experiments, we repeat the experiment $20$ times and take the average to get stable results.}. We also include two non-contextual word representations -- GloVe and ELMo-0 as baselines. We report $MIG$ as the results of our probe on the two graph structures in Figures~\ref{fig:mi_layers_ptb} and ~\ref{fig:mi_layers_amr}.

The $MIG$ estimations for syntactic structure probing of both BERT-base and BERT-large are quite high, which implies that BERT encodes much syntactic information. However, for the semantic structure, the $MIG$ scores of BERT models are lower, suggesting that BERT does not encode the semantic structures as well. These two conclusions are consistent with previous works~\citep{bert_no_semantic, edge_probe, wu2020infusing} which have found that unlike syntax, semantics is not captured well by the pretrained models.

We also observe an interesting trend when comparing $MIG$ across layers. We find that for syntax, $MIG$ starts to decrease in the upper layers, especially for the BERT-large. 
This is consistent with previous works which report that BERT models syntax more in the lower and middle layers~\citep{tenney-etal-2019-bert}. 
For semantic graphs, $MIG$ is steady across all layers. It means that semantic information is spread across the entire model. The results are consistent with existing work~\citep{primer_bert_RogersKR20}.
For the two non-contextual baselines, GloVe and ELMo-0, we can see that their $MIG$ scores are lower compared with contextualized representations, especially for syntax. Previous work~\citep{structural_probe} has drawn similar conclusions. While for the semantic graphs, the gap is not significant.


\subsection{Probing Localized Information}
In this section, we show how we can use the \we probe to understand if the contextualized representations capture localized linguistic information in the dependency parses such as POS information or relational dependency information.
As described before, we design various perturbation experiments using our \we probe. For probing POS information or a dependency relation type, we add noise to the graph embeddings of the corresponding node(s).
After that, we calculate the $MIL$ ratio (eq.~\ref{eq:ratio_local}) to show how much particular linguistic information (POS or relation type information) is contained in the word representations. We repeat the experiment $20$ times and use boxplots to present all the results. 


\begin{figure}[ht]
	\centering
	\includegraphics[scale=0.25]{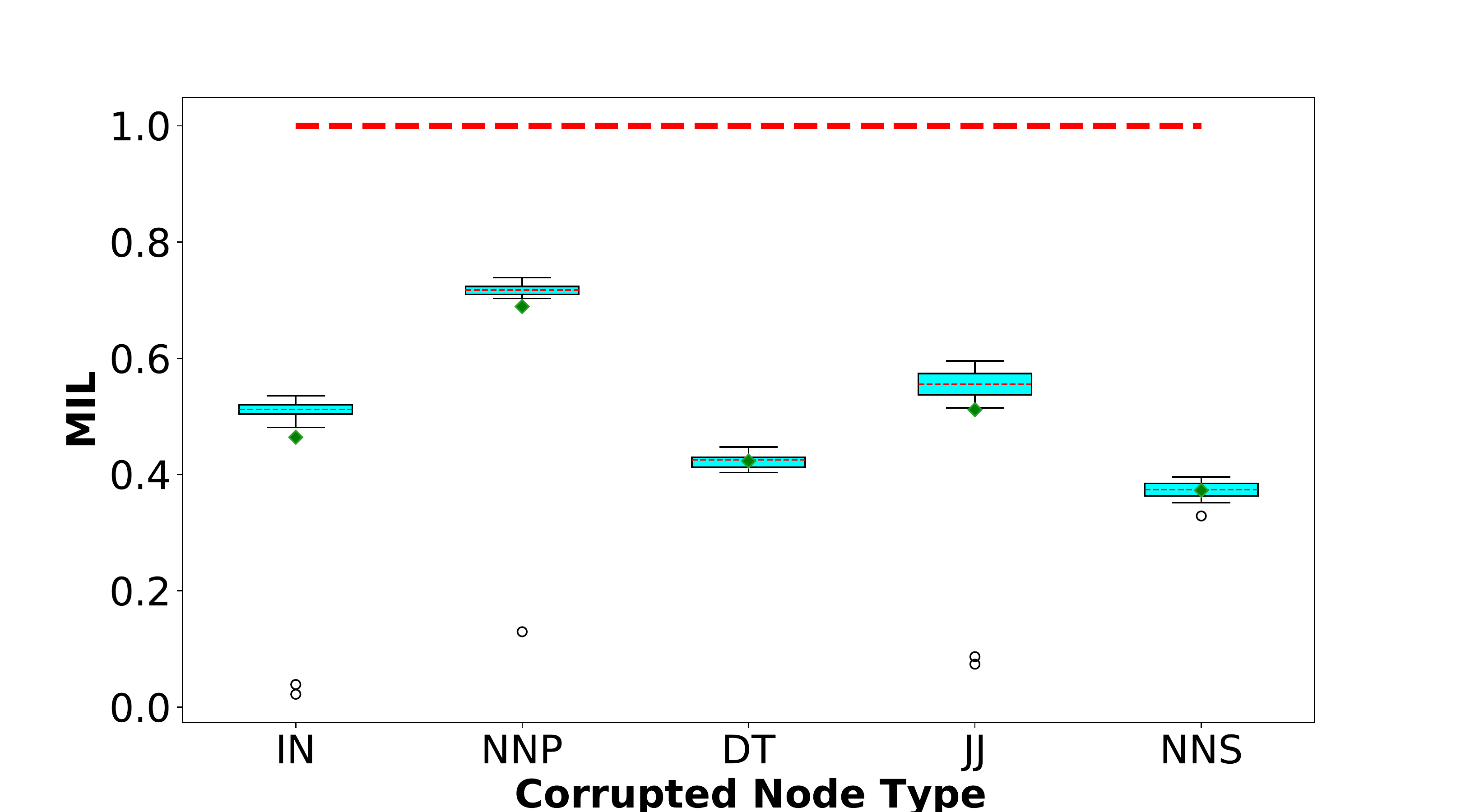}
	\vspace{-6.mm}
	\caption{$MIL$ scores of probing $5$ types of POS tags (localized syntactic structure) for word representations in BERT-base (output layer). The local structure is decided by the POS tags attached on nodes.}
	\label{fig:noisy_pos}
	\vspace{-3.mm}
\end{figure}
First we use \we to test for POS information, which is tagged as node labels in the dependency tree. We select $5$ POS tags: \textit{IN}, \textit{NNP}, \textit{DT}, \textit{JJ}, and \textit{NNS}, which have high and roughly the same frequencies in the Penn Treebank dataset.
Complete statistics about the POS tag frequencies can be found in Appendix~\ref{appendix:ptb}. We ensure that the amount of perturbation of the graph embeddings is the same for each type. 
Figure~\ref{fig:noisy_pos} presents the results. We find that \textit{NNP} achieves the highest $MIL$ score, while \textit{NNS} achieves the lowest. This implies that BERT encodes syntactic information for singular proper nouns (\textit{NNP}) and adjectives (\textit{JJ}) more than plural nouns (\textit{NNS}).

\begin{figure}[ht]
	\centering
	\includegraphics[scale=0.25]{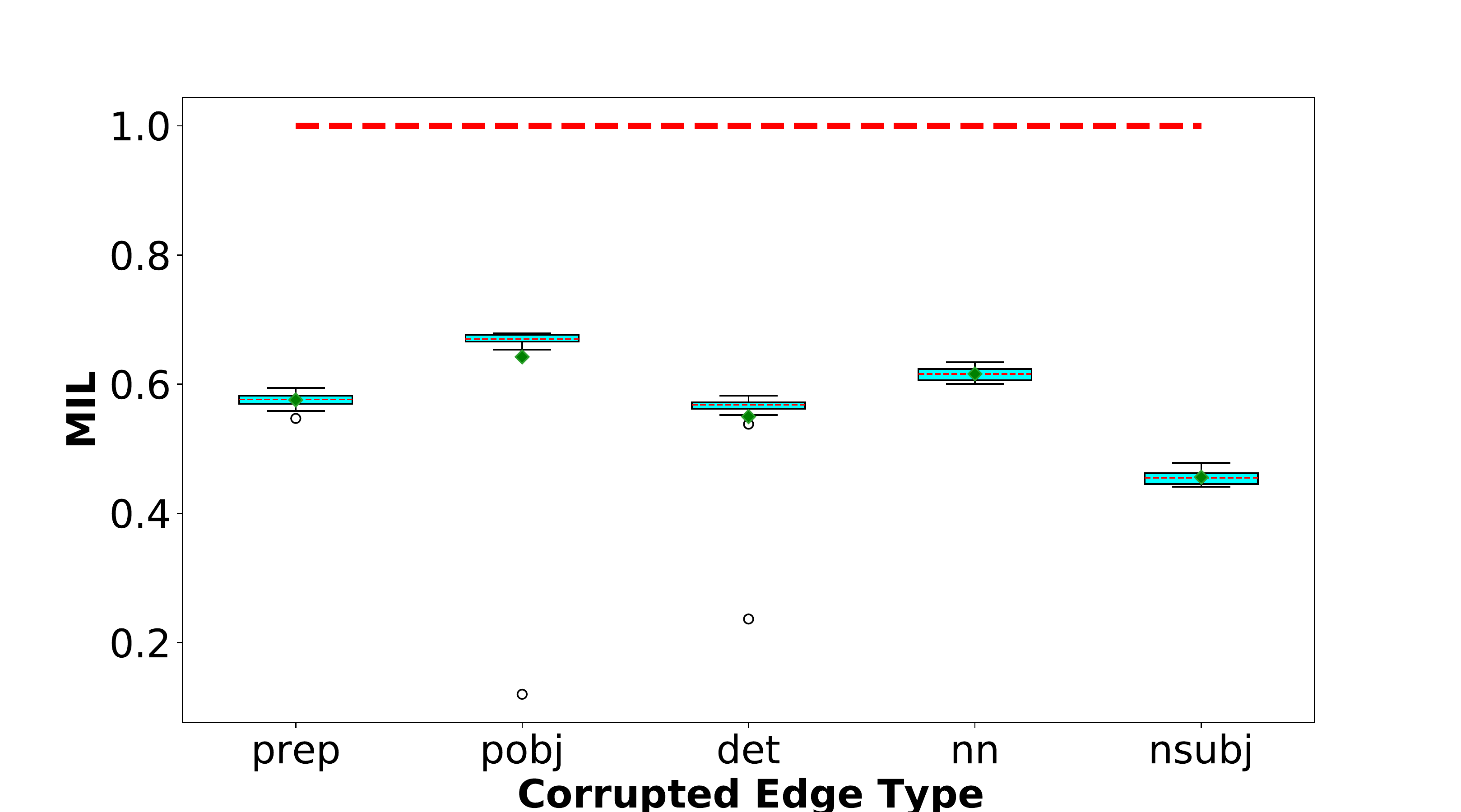}
	\vspace{-6.mm}
	\caption{$MIL$ scores of probing $5$ types of localized syntactic structure for word representations in BERT-base (output layer). The local structure is decided by the dependency relation labels attached on edges.}
	\label{fig:noisy_ptb}
	\vspace{-3.mm}
\end{figure}
Next, we probe $5$ types of universal dependency relations in the Penn Treebank dataset (PTB). These are \textit{prep}, \textit{pobj}, \textit{det}, \textit{nn} and \textit{nsubj}. These $5$ relations also roughly occur the same number of times in PTB. Complete statistics about the number of occurences of these relation types can be found in Appendix~\ref{appendix:ptb}. Similarly, for each type of relation, we add same amount of perturbation to graph embeddings of nodes connected by the specific relations. 
Figure~\ref{fig:noisy_ptb} shows the results, where \textit{nsubj} relations have the lowest $MIL$ score compared with other $4$ types. This means that BERT encodes more syntactic structure for prepositional modifiers (\textit{prep}), object of a preposition (\textit{pobj}), and noun compound modifier (\textit{nn}) than nominal subject (\textit{nsubj}). \citet{emily_geometric_bert} have drawn similar conclusions while probing for dependency arc labels. 
Similar experiment for semantic structure can be found in Appendix~\ref{appendix:semantic_local}.

\begin{figure}[tbp]
	\centering
	\includegraphics[scale=0.25]{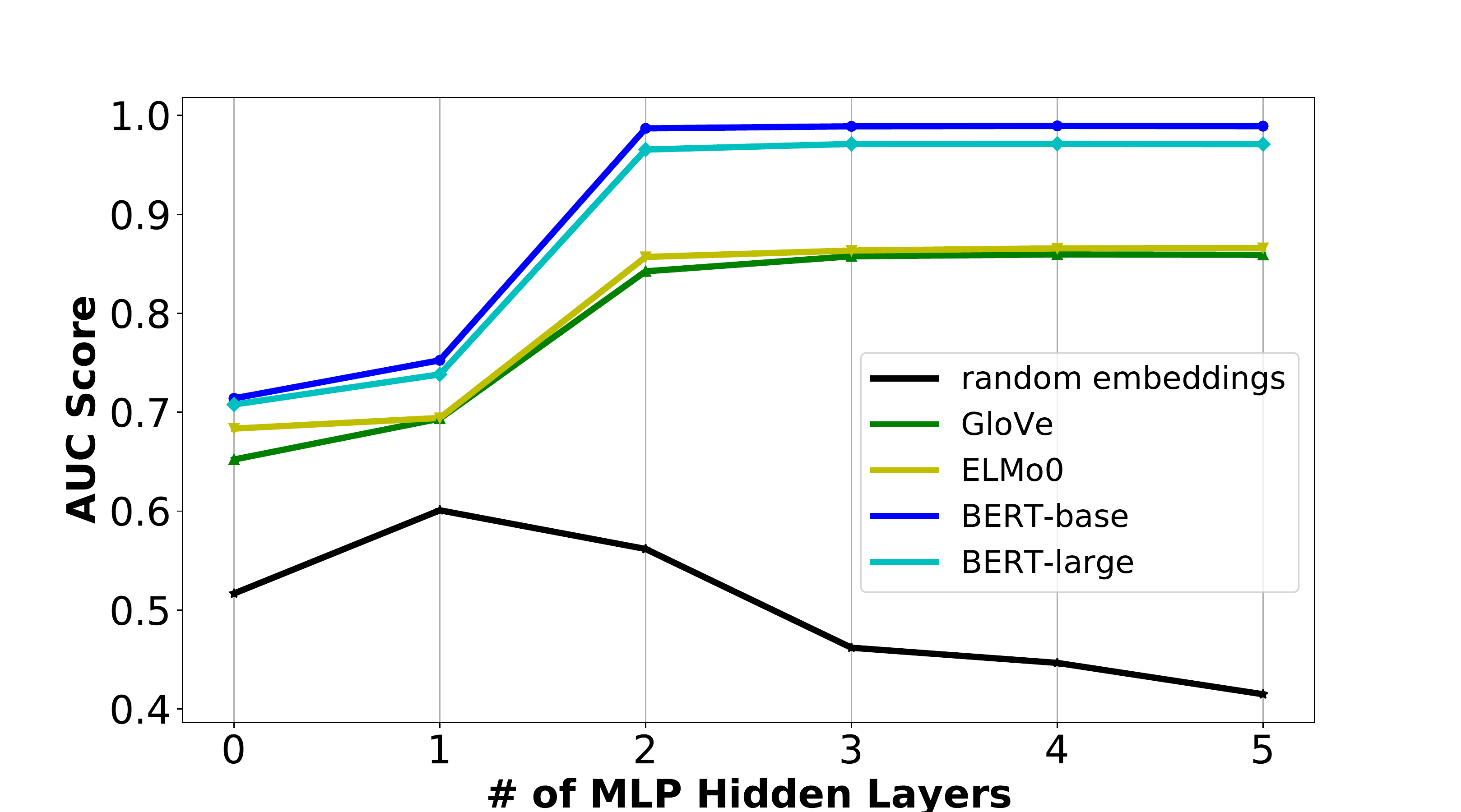}
	\vspace{-6.mm}
	\caption{AUC scores of predicting syntactic trees by various word representations.}
	\vspace{-3.mm}
	\label{fig:graph_test_ptb}
\end{figure}
\begin{figure}[tbp]
	\centering
	\includegraphics[scale=0.25]{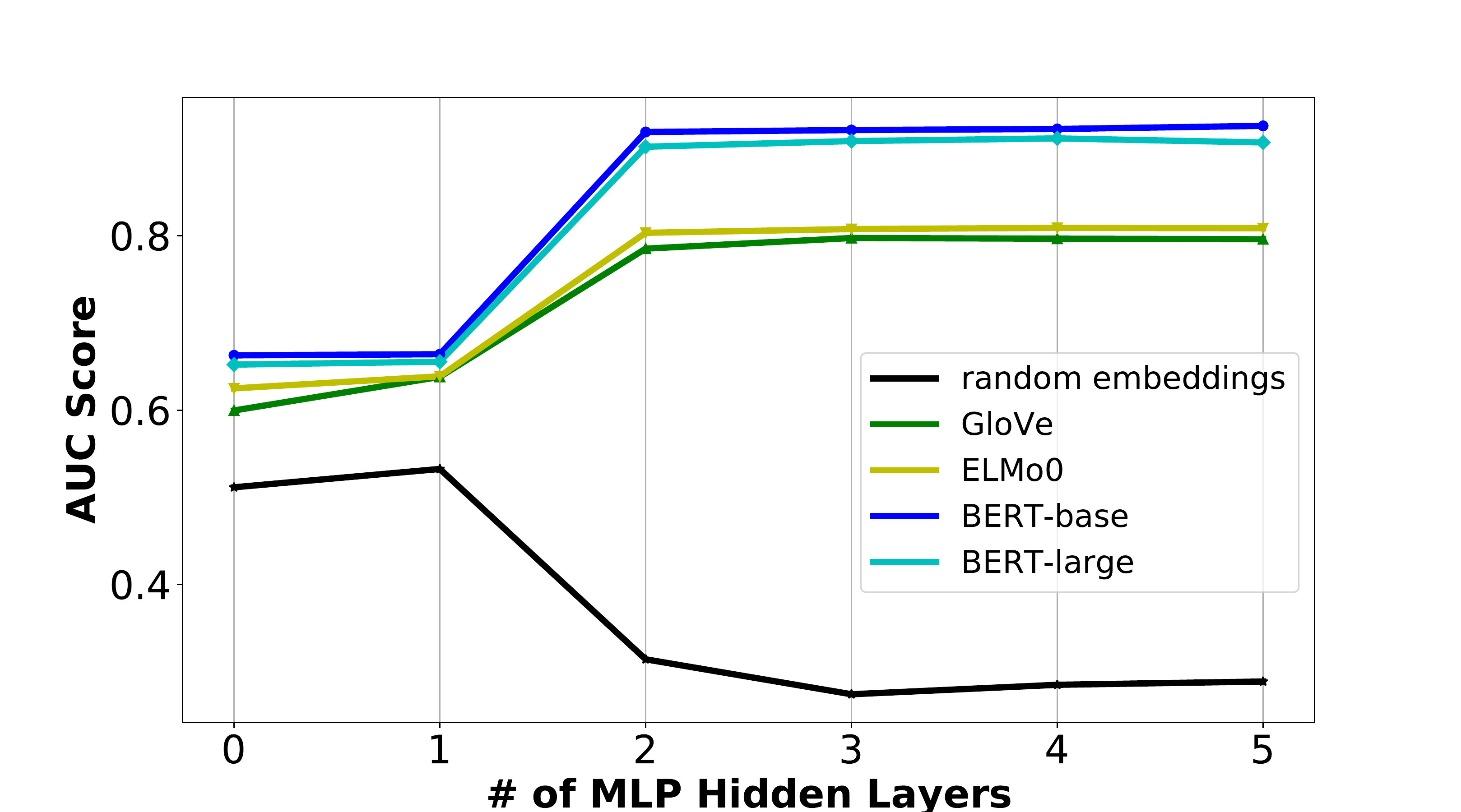}
	\vspace{-6.mm}
	\caption{AUC scores of predicting semantic graphs by various word representations.}
	\vspace{-3.mm}
	\label{fig:graph_test_amr}
\end{figure}

\subsection{On Accuracy-Based Probing}
In contrast to our information-theoretic approach to probing, we train a group of MLP models to probe entire and local structures in BERT-base. We show that these probe results mainly depend on the model complexity rather than the structure itself.

\textbf{Probing entire graph structures.} A group of MLPs are trained to predict entire syntactic and semantic structures with word representations. 
Figure~\ref{fig:graph_test_ptb} and Figure~\ref{fig:graph_test_amr} show the results. Their trends are similar.
Shallow MLPs perform the worst and deep ones perform much better.
Previous work on structural probing~\citep{structural_probe} argues that powerful models could parse the word representations, thus a simple model should be designed. However, in Table~\ref{tab:ge_test}, we find that linear model even could not restore the graph by its embeddings. Obviously, its performance cannot indicate how much structure information is included in the graph embeddings. Thus, there is no reasonable principle to decide the complexity of the probe model. Given this, designing and training a model is not suitable to probe entire structures. A similar argument has been placed by previous works~\citep{pimentel-etal-2020-pareto,mi_probe,lovering2021informationtheoretic}. 


\begin{table}[ht]
	\caption{Performance (AUC score) of predicting graph structures with MLPs of different complexity}
	\label{tab:model_complexity}
	\vspace{-2mm}
	\scalebox{0.65}{
		\begin{tabular}{c|c|c|c|c|c|c}
			\hline
			\toprule
			Relation Type & linear & 1 & 2 & 3 & 4 & 5 \\
			\midrule
			\hline
			prep & 0.6966 & \textbf{0.7232} & 0.9838 & 0.9858 & 0.9863 & 0.9866 \\
            pobj & 0.6224 & 0.6619 & 0.9874 & 0.9888 & 0.9891 & 0.9894 \\
            det & \textbf{0.7016} & 0.7222 & \textbf{0.9928} & \textbf{0.9938} & \textbf{0.9940} & \textbf{0.9943} \\
            nn & 0.6877 & 0.7221 & 0.9884 & 0.9899 & 0.9904 & 0.9903 \\
            nsubj & 0.6754 & 0.6938 & 0.9841 & 0.9859 & 0.9868 & 0.9869 \\
			\hline
			\hline
			arg & \textbf{0.6686} & \textbf{0.6652} & 0.9185 & 0.9217 & 0.9189 & 0.9215 \\
			general & 0.6621 & 0.6574 & \textbf{0.9192} & \textbf{0.9223} & \textbf{0.9221} & \textbf{0.9221} \\
            op & 0.6500 & 0.6500 & 0.9098 & 0.9177 & 0.9130 & 0.9160 \\
			\bottomrule
			\hline
		\end{tabular}}
		\vspace{-2.mm}
\end{table}
\textbf{Probing localized graph structures.} To prove that accuracies of probe models for localized structure also mainly depend on the model's complexity rather than the local structure, we train the group of MLPs to predict the entire syntactic structure by word representations, and calculate the AUC scores for each type of relations in test set as probing results. Table~\ref{tab:model_complexity} shows the AUC score of predicting specific type of relations. For syntactic structure, same $5$ types of relations are selected, and for semantic graphs, we select $3$ groups of relations to probe: \textit{arg}, \textit{general} and \textit{op}. Complete statistics of AMR Bank are in Appendix~\ref{appendix:ptb}.
From the results, we can find that for MLP models with different number of hidden layers, the ranks of AUC scores of relation prediction are quite different.
For both syntax trees and semantic graphs, there is no consistent interpretation of the results to conclude which types of relations are encoded in BERT. We also run the experiment in the perturbation settings, which can be referred to Appendix~\ref{appendix:perturbation_accuracy}.

Combining the results of probing with accuracy in Figure~\ref{fig:graph_test_ptb}, Figure~\ref{fig:graph_test_amr}, and Table~\ref{tab:model_complexity}, we can find that the prediction decisions are not based purely on the structure but rather on spurious heuristics. This has also been concluded and discussed in some recent works~\citep{hewitt_control_task, lovering2021informationtheoretic}.
Thus, training models is not feasible to probe structures.
For our probe methods, the randomness of models such as complexity is not an issue, since the one with highest estimation should be selected for tighter compression lemma bound~\ref{eq:kl_bound} as introduced by~\citet{mi_probe}.

\subsection{Hyperparameter and Efficiency}
Information-theoretic approaches sacrifice simplicity and efficiency to achieve reliable probing results compared to accuracy-based probes. Even though our probes are quite simple, there are more hyperparameters that need to be selected by users compared to accuracy-based probes. To help users implement our methods in their setting, we briefly describe some guiding principles to help them select hyperparameters, and point out several potential ways to make our probing approach more efficient.

Our probes are composed of two steps: (a) computation of the graph embedding, and (b) estimation of the mutual information. The guiding principal in the graph embedding step is to retain as much linguistic graph information as possible. In our experiments, we used default hyperparameters in DeepWalk~\citep{deepwalk} for simplicity. Details can be found in Appendix~\ref{appendix:ge}. However, users may use also use other graph embedding approaches that incorporate edge labels, etc. to improve our model. As the mutual information estimation procedure is estimating a lower bound to the true mutual information, the guiding principle for hyperparameter selection in this step should be to let the MI estimation values be as large as possible. In particular, model size is worth noting. Deeper models can achieve a tighter lower bound. However, these are less efficient than shallow ones. 
Thus, the selection of MI estimator's complexity is a tradeoff. According to our empirical experience, a relatively good choice is to use a two-layer MLPs. More details can be found in Appendix~\ref{appendix:mi_estimation}.
Note that it might also be harder to achieve convergence with deeper models as training of MI estimators is notoriously difficult. We leave a better exploration of this to future work.

Potential users might also resort to other solutions to make the probes more efficient.
If the bottleneck is in the graph embedding step, some fast approaches~\citep{graphsage_HamiltonYL17,line_TangQWZYM15} can be chosen instead. If the mutual information estimation step is the bottlenneck, some sampling strategies can be used. A simple way is to sample a subset of the dataset, and optimize eq.~\ref{eq:obj} based on that subset. Alternatively, potential users can use more sophisticated sampling strategies in training as in ~\newcite{random_sample_training_recht2012beneath}. These approaches achieve a much better convergence rate for MI estimation.

\section{Related Work}
\textbf{Syntax and Semantics Probing.}
Many existing works probe language models directly or indirectly showing how much syntactic and semantic information is encoded in them. \citet{pos_probe} tested NMT models and found that higher layers encode semantic information while lower layers perform better at POS tagging. Similarly, \citet{bert_probe} tested various BERT layers and found that it encodes a rich hierarchy of linguistic information in the intermediate
layers. 
\citet{edge_probe, wu2020infusing} compared the syntactic and semantic information in BERT and its variants, and found that more syntactic information is encoded than semantic information.
\citet{morphology_probe} focused on probing various linguistic features with 10 different designed tasks. \citet{structural_probe} designed a tree distance and depth prediction task to probe syntax tree structures. 

\textbf{Information Theoretic Probe.}
With the popularity of probe methods, limitations of previous methods have also been found. Information theoretic methods have been proposed as an alternative.
To avoid the randomness of performance brought by the varying sizes of the probe models, \citet{mi_probe} proposed an information-theoretic probe with control functions, which used mutual information instead of model performance for probing. \citet{mdl_probe} restricted the probe model size by Minimum Description Length. Training a model is recast as teaching it to effectively transmit the data. \citet{lovering2021informationtheoretic} pointed out that if we train a model to probe, the decisions are often not based on information itself, but rather on spurious heuristics specific to the training set.

\textbf{Mutual Information Estimation.}
Mutual information estimation is a well-known difficult problem, especially when the feature vectors are in a high dimensional space~\citep{chow_mi_estimation, peng_mi_estimation}. There are many traditional ways to estimate MI, such as the well-known histogram approach~\citep{steuer_mi_estimation_histogram, paninski_mi_estimation_histogram}, density estimations using a kernel~\citep{moon1995_mi_estimation_kernel}, and nearest-neighbor distance~\citep{kraskov2004_mi_estimation_nn}. \citet{mine} was recently proposed as a way to estimate MI using neural networks, which showed marked improvement over previous methods for feature vectors in high-dimensional space.

\section{Limitations and Future Work}
In this paper we propose a general information-theoretic probe method, which is capable of probing for linguistic graph structures and avoids the randomness of training a model. In the experiments, we use our probe method to show the extent to which syntax trees and semantic graphs are encoded in pretrained BERT models. Further, we perform a simple perturbation analysis to show that with small modifications, the probe can also be used to probe for specific linguistic sub-structures.
There are some limitations of our probe. First, a graph embedding is used, and some structure information could be lost in this process. We provide simple ways to test this. Second, training a MI estimation model is difficult.
Future work can consider building on our framework by exploring better graph embedding and MI estimation techniques.


\section*{Broader Impact and Discussion of Ethics}
In recent years, deep learning approaches have been the main models for state-of-the-art systems in natural language processing. However, understanding the decision making in these systems has been hard, and has challenges when these systems are used in human contexts.
Probing helps us gain interpretability and hence is useful in deploying these black-box models. Our work introduces a simple and general way for understanding how linguistic properties represented as graph structures are encoded in large pretrained language models which are being applied to a wide range of structures in NLP. The methodology and probing results can be helpful to the development of future NLP models.

While our model is not tuned for any specific real-world application domain, our methods could be used in sensitive contexts such as legal or health-care settings, and it is essential that any work using our probe method undertake extensive quality-assurance and robustness testing before using it in their setting. The datasets used in our work do not contain any sensitive information to the best of our knowledge.

\section*{Acknowledgments}
We would like to thank reviewers for the constructive comments and providing suggestions for future work. This work was funded by SNF project \#201009.

\bibliographystyle{acl_natbib}
\bibliography{acl2021}

\begin{thebibliography}{49}
\expandafter\ifx\csname natexlab\endcsname\relax\def\natexlab#1{#1}\fi

\bibitem[{Abend and Rappoport(2013)}]{abend2013universal}
Omri Abend and Ari Rappoport. 2013.
\newblock \href {https://www.aclweb.org/anthology/P13-1023/} {Universal
  conceptual cognitive annotation {(UCCA)}}.
\newblock In \emph{Proceedings of the 51st Annual Meeting of the Association
  for Computational Linguistics, {ACL} 2013, 4-9 August 2013, Sofia, Bulgaria,
  Volume 1: Long Papers}, pages 228--238. The Association for Computer
  Linguistics.

\bibitem[{Banarescu et~al.(2013)Banarescu, Bonial, Cai, Georgescu, Griffitt,
  Hermjakob, Knight, Koehn, Palmer, and Schneider}]{AMR}
Laura Banarescu, Claire Bonial, Shu Cai, Madalina Georgescu, Kira Griffitt, Ulf
  Hermjakob, Kevin Knight, Philipp Koehn, Martha Palmer, and Nathan Schneider.
  2013.
\newblock \href {https://www.aclweb.org/anthology/W13-2322/} {Abstract meaning
  representation for sembanking}.
\newblock In \emph{Proceedings of the 7th Linguistic Annotation Workshop and
  Interoperability with Discourse, LAW-ID@ACL 2013, August 8-9, 2013, Sofia,
  Bulgaria}, pages 178--186. The Association for Computer Linguistics.

\bibitem[{Banerjee(2006)}]{bound}
Arindam Banerjee. 2006.
\newblock \href {https://doi.org/10.1145/1143844.1143855} {On bayesian bounds}.
\newblock In \emph{Machine Learning, Proceedings of the Twenty-Third
  International Conference {(ICML} 2006), Pittsburgh, Pennsylvania, USA, June
  25-29, 2006}, volume 148 of \emph{{ACM} International Conference Proceeding
  Series}, pages 81--88. {ACM}.

\bibitem[{Belghazi et~al.(2018)Belghazi, Baratin, Rajeswar, Ozair, Bengio,
  Hjelm, and Courville}]{mine}
Mohamed~Ishmael Belghazi, Aristide Baratin, Sai Rajeswar, Sherjil Ozair, Yoshua
  Bengio, R.~Devon Hjelm, and Aaron~C. Courville. 2018.
\newblock \href {http://proceedings.mlr.press/v80/belghazi18a.html} {Mutual
  information neural estimation}.
\newblock In \emph{Proceedings of the 35th International Conference on Machine
  Learning, {ICML} 2018, Stockholmsm{\"{a}}ssan, Stockholm, Sweden, July 10-15,
  2018}, volume~80 of \emph{Proceedings of Machine Learning Research}, pages
  530--539. {PMLR}.

\bibitem[{Belinkov et~al.(2017)Belinkov, M{\`{a}}rquez, Sajjad, Durrani, Dalvi,
  and Glass}]{pos_probe}
Yonatan Belinkov, Llu{\'{\i}}s M{\`{a}}rquez, Hassan Sajjad, Nadir Durrani,
  Fahim Dalvi, and James~R. Glass. 2017.
\newblock \href {https://www.aclweb.org/anthology/I17-1001/} {Evaluating layers
  of representation in neural machine translation on part-of-speech and
  semantic tagging tasks}.
\newblock In \emph{Proceedings of the Eighth International Joint Conference on
  Natural Language Processing, {IJCNLP} 2017, Taipei, Taiwan, November 27 -
  December 1, 2017 - Volume 1: Long Papers}, pages 1--10. Asian Federation of
  Natural Language Processing.

\bibitem[{Bos(2013)}]{bos2017groningen}
Johan Bos. 2013.
\newblock \href {https://www.aclweb.org/anthology/W13-3802/} {The groningen
  meaning bank}.
\newblock In \emph{Proceedings of the Joint Symposium on Semantic Processing.
  Textual Inference and Structures in Corpora, {JSSP} 2013, Trento, Italy,
  November 20-22, 2013}, page~2. Association for Computational Linguistics.

\bibitem[{Cai et~al.(2018)Cai, Zheng, and Chang}]{hongyun_ge_survey}
Hongyun Cai, Vincent~W. Zheng, and Kevin~Chen{-}Chuan Chang. 2018.
\newblock \href {https://doi.org/10.1109/TKDE.2018.2807452} {A comprehensive
  survey of graph embedding: Problems, techniques, and applications}.
\newblock \emph{{IEEE} Trans. Knowl. Data Eng.}, 30(9):1616--1637.

\bibitem[{Chomsky(1957)}]{chomsky1957logical}
Noam Chomsky. 1957.
\newblock Logical structures in language.
\newblock \emph{American Documentation (pre-1986)}, 8(4):284.

\bibitem[{Chow and Huang(2005)}]{chow_mi_estimation}
Tommy W.~S. Chow and Di~Huang. 2005.
\newblock \href {https://doi.org/10.1109/TNN.2004.841414} {Estimating optimal
  feature subsets using efficient estimation of high-dimensional mutual
  information}.
\newblock \emph{{IEEE} Trans. Neural Networks}, 16(1):213--224.

\bibitem[{Conneau et~al.(2018)Conneau, Kruszewski, Lample, Barrault, and
  Baroni}]{morphology_probe}
Alexis Conneau, Germ{\'{a}}n Kruszewski, Guillaume Lample, Lo{\"{\i}}c
  Barrault, and Marco Baroni. 2018.
\newblock \href {https://doi.org/10.18653/v1/P18-1198} {What you can cram into
  a single {\textbackslash}{\textdollar}{\&}!{\#}* vector: Probing sentence
  embeddings for linguistic properties}.
\newblock In \emph{Proceedings of the 56th Annual Meeting of the Association
  for Computational Linguistics, {ACL} 2018, Melbourne, Australia, July 15-20,
  2018, Volume 1: Long Papers}, pages 2126--2136. Association for Computational
  Linguistics.

\bibitem[{Cristiani et~al.(2020)Cristiani, Lecomte, and
  Maurine}]{mi_estimation_loss}
Valence Cristiani, Maxime Lecomte, and Philippe Maurine. 2020.
\newblock \href {https://doi.org/10.1007/978-3-030-61638-0\_9} {Leakage
  assessment through neural estimation of the mutual information}.
\newblock In \emph{Applied Cryptography and Network Security Workshops - {ACNS}
  2020 Satellite Workshops, AIBlock, AIHWS, AIoTS, Cloud S{\&}P, SCI, SecMT,
  and SiMLA, Rome, Italy, October 19-22, 2020, Proceedings}, volume 12418 of
  \emph{Lecture Notes in Computer Science}, pages 144--162. Springer.

\bibitem[{De~Marneffe et~al.(2006)De~Marneffe, MacCartney, Manning
  et~al.}]{Marneffe:2006}
Marie-Catherine De~Marneffe, Bill MacCartney, Christopher~D Manning, et~al.
  2006.
\newblock Generating typed dependency parses from phrase structure parses.
\newblock In \emph{Lrec}, volume~6, pages 449--454.

\bibitem[{Devlin et~al.(2019)Devlin, Chang, Lee, and Toutanova}]{bert}
Jacob Devlin, Ming{-}Wei Chang, Kenton Lee, and Kristina Toutanova. 2019.
\newblock \href {https://doi.org/10.18653/v1/n19-1423} {{BERT:} pre-training of
  deep bidirectional transformers for language understanding}.
\newblock In \emph{Proceedings of the 2019 Conference of the North American
  Chapter of the Association for Computational Linguistics: Human Language
  Technologies, {NAACL-HLT} 2019, Minneapolis, MN, USA, June 2-7, 2019, Volume
  1 (Long and Short Papers)}, pages 4171--4186. Association for Computational
  Linguistics.

\bibitem[{Escolano et~al.(2017)Escolano, Hancock, Lozano, and
  Curado}]{francisco_mi_graphs}
Francisco Escolano, Edwin~R. Hancock, Miguel~Angel Lozano, and Manuel Curado.
  2017.
\newblock \href {https://doi.org/10.1016/j.patrec.2016.07.012} {The mutual
  information between graphs}.
\newblock \emph{Pattern Recognit. Lett.}, 87:12--19.

\bibitem[{Fawcett(2006)}]{tom_auc}
Tom Fawcett. 2006.
\newblock \href {https://doi.org/10.1016/j.patrec.2005.10.010} {An introduction
  to {ROC} analysis}.
\newblock \emph{Pattern Recognit. Lett.}, 27(8):861--874.

\bibitem[{Gao et~al.(2015)Gao, Steeg, and Galstyan}]{gao_mi_scale}
Shuyang Gao, Greg~Ver Steeg, and Aram Galstyan. 2015.
\newblock \href {http://proceedings.mlr.press/v38/gao15.html} {Efficient
  estimation of mutual information for strongly dependent variables}.
\newblock In \emph{Proceedings of the Eighteenth International Conference on
  Artificial Intelligence and Statistics, {AISTATS} 2015, San Diego,
  California, USA, May 9-12, 2015}, volume~38 of \emph{{JMLR} Workshop and
  Conference Proceedings}. JMLR.org.

\bibitem[{Hajic et~al.(2012)Hajic, Hajicov{\'{a}}, Panevov{\'{a}}, Sgall,
  Bojar, Cinkov{\'{a}}, Fuc{\'{\i}}kov{\'{a}}, Mikulov{\'{a}}, Pajas, Popelka,
  Semeck{\'{y}}, Sindlerov{\'{a}}, Step{\'{a}}nek, Toman, Uresov{\'{a}}, and
  Zabokrtsk{\'{y}}}]{hajic-etal-2012-announcing}
Jan Hajic, Eva Hajicov{\'{a}}, Jarmila Panevov{\'{a}}, Petr Sgall, Ondrej
  Bojar, Silvie Cinkov{\'{a}}, Eva Fuc{\'{\i}}kov{\'{a}}, Marie Mikulov{\'{a}},
  Petr Pajas, Jan Popelka, Jir{\'{\i}} Semeck{\'{y}}, Jana Sindlerov{\'{a}},
  Jan Step{\'{a}}nek, Josef Toman, Zdenka Uresov{\'{a}}, and Zdenek
  Zabokrtsk{\'{y}}. 2012.
\newblock \href
  {http://www.lrec-conf.org/proceedings/lrec2012/summaries/510.html}
  {Announcing prague czech-english dependency treebank 2.0}.
\newblock In \emph{Proceedings of the Eighth International Conference on
  Language Resources and Evaluation, {LREC} 2012, Istanbul, Turkey, May 23-25,
  2012}, pages 3153--3160. European Language Resources Association {(ELRA)}.

\bibitem[{Hamilton et~al.(2017)Hamilton, Ying, and
  Leskovec}]{graphsage_HamiltonYL17}
William~L. Hamilton, Zhitao Ying, and Jure Leskovec. 2017.
\newblock \href
  {https://proceedings.neurips.cc/paper/2017/hash/5dd9db5e033da9c6fb5ba83c7a7ebea9-Abstract.html}
  {Inductive representation learning on large graphs}.
\newblock In \emph{Advances in Neural Information Processing Systems 30: Annual
  Conference on Neural Information Processing Systems 2017, December 4-9, 2017,
  Long Beach, CA, {USA}}, pages 1024--1034.

\bibitem[{Hewitt and Liang(2019)}]{hewitt_control_task}
John Hewitt and Percy Liang. 2019.
\newblock \href {https://doi.org/10.18653/v1/D19-1275} {Designing and
  interpreting probes with control tasks}.
\newblock In \emph{Proceedings of the 2019 Conference on Empirical Methods in
  Natural Language Processing and the 9th International Joint Conference on
  Natural Language Processing, {EMNLP-IJCNLP} 2019, Hong Kong, China, November
  3-7, 2019}, pages 2733--2743. Association for Computational Linguistics.

\bibitem[{Hewitt and Manning(2019)}]{structural_probe}
John Hewitt and Christopher~D. Manning. 2019.
\newblock \href {https://doi.org/10.18653/v1/n19-1419} {A structural probe for
  finding syntax in word representations}.
\newblock In \emph{Proceedings of the 2019 Conference of the North American
  Chapter of the Association for Computational Linguistics: Human Language
  Technologies, {NAACL-HLT} 2019, Minneapolis, MN, USA, June 2-7, 2019, Volume
  1 (Long and Short Papers)}, pages 4129--4138. Association for Computational
  Linguistics.

\bibitem[{Hockenmaier and Steedman(2007)}]{hockenmaier2007ccgbank}
Julia Hockenmaier and Mark Steedman. 2007.
\newblock \href {https://doi.org/10.1162/coli.2007.33.3.355} {Ccgbank: {A}
  corpus of {CCG} derivations and dependency structures extracted from the penn
  treebank}.
\newblock \emph{Comput. Linguistics}, 33(3):355--396.

\bibitem[{Huang et~al.(1993)Huang, Alleva, Hon, Hwang, Lee, and
  Rosenfeld}]{skipgram}
Xuedong Huang, Fileno~A. Alleva, Hsiao{-}Wuen Hon, Mei{-}Yuh Hwang, Kai{-}Fu
  Lee, and Ronald Rosenfeld. 1993.
\newblock \href {https://doi.org/10.1006/csla.1993.1007} {The {SPHINX-II}
  speech recognition system: an overview}.
\newblock \emph{Comput. Speech Lang.}, 7(2):137--148.

\bibitem[{Jawahar et~al.(2019)Jawahar, Sagot, and Seddah}]{bert_probe}
Ganesh Jawahar, Beno{\^{\i}}t Sagot, and Djam{\'{e}} Seddah. 2019.
\newblock \href {https://doi.org/10.18653/v1/p19-1356} {What does {BERT} learn
  about the structure of language?}
\newblock In \emph{Proceedings of the 57th Conference of the Association for
  Computational Linguistics, {ACL} 2019, Florence, Italy, July 28- August 2,
  2019, Volume 1: Long Papers}, pages 3651--3657. Association for Computational
  Linguistics.

\bibitem[{Koller et~al.(2019)Koller, Oepen, and Sun}]{koller2019graph}
Alexander Koller, Stephan Oepen, and Weiwei Sun. 2019.
\newblock \href {https://doi.org/10.18653/v1/p19-4002} {Graph-based meaning
  representations: Design and processing}.
\newblock In \emph{Proceedings of the 57th Conference of the Association for
  Computational Linguistics: Tutorial Abstracts, {ACL} 2019, Florence, Italy,
  July 28, 2019, Volume 4: Tutorial Abstracts}, pages 6--11. Association for
  Computational Linguistics.

\bibitem[{Kraskov et~al.(2004)Kraskov, St{\"o}gbauer, and
  Grassberger}]{kraskov2004_mi_estimation_nn}
Alexander Kraskov, Harald St{\"o}gbauer, and Peter Grassberger. 2004.
\newblock Estimating mutual information.
\newblock \emph{Physical review E}, 69(6):066138.

\bibitem[{Liu et~al.(2019)Liu, Gardner, Belinkov, Peters, and
  Smith}]{bert_no_semantic}
Nelson~F. Liu, Matt Gardner, Yonatan Belinkov, Matthew~E. Peters, and Noah~A.
  Smith. 2019.
\newblock \href {https://doi.org/10.18653/v1/n19-1112} {Linguistic knowledge
  and transferability of contextual representations}.
\newblock In \emph{Proceedings of the 2019 Conference of the North American
  Chapter of the Association for Computational Linguistics: Human Language
  Technologies, {NAACL-HLT} 2019, Minneapolis, MN, USA, June 2-7, 2019, Volume
  1 (Long and Short Papers)}, pages 1073--1094. Association for Computational
  Linguistics.

\bibitem[{Lovering et~al.(2021)Lovering, Jha, Linzen, and
  Pavlick}]{lovering2021informationtheoretic}
Charles Lovering, Rohan Jha, Tal Linzen, and Ellie Pavlick. 2021.
\newblock \href {https://openreview.net/forum?id=mNtmhaDkAr} {Predicting
  inductive biases of pre-trained models}.
\newblock In \emph{International Conference on Learning Representations}.

\bibitem[{Marcus et~al.(1993)Marcus, Santorini, and
  Marcinkiewicz}]{dataset_ptb}
Mitchell~P. Marcus, Beatrice Santorini, and Mary~Ann Marcinkiewicz. 1993.
\newblock Building a large annotated corpus of english: The penn treebank.
\newblock \emph{Comput. Linguistics}, 19(2):313--330.

\bibitem[{de~Marneffe et~al.(2006)de~Marneffe, MacCartney, and Manning}]{SDF}
Marie{-}Catherine de~Marneffe, Bill MacCartney, and Christopher~D. Manning.
  2006.
\newblock \href
  {http://www.lrec-conf.org/proceedings/lrec2006/summaries/440.html}
  {Generating typed dependency parses from phrase structure parses}.
\newblock In \emph{Proceedings of the Fifth International Conference on
  Language Resources and Evaluation, {LREC} 2006, Genoa, Italy, May 22-28,
  2006}, pages 449--454. European Language Resources Association {(ELRA)}.

\bibitem[{Mikolov et~al.(2013)Mikolov, Sutskever, Chen, Corrado, and
  Dean}]{word2vec}
Tom{\'{a}}s Mikolov, Ilya Sutskever, Kai Chen, Gregory~S. Corrado, and Jeffrey
  Dean. 2013.
\newblock \href
  {https://proceedings.neurips.cc/paper/2013/hash/9aa42b31882ec039965f3c4923ce901b-Abstract.html}
  {Distributed representations of words and phrases and their
  compositionality}.
\newblock In \emph{Advances in Neural Information Processing Systems 26: 27th
  Annual Conference on Neural Information Processing Systems 2013. Proceedings
  of a meeting held December 5-8, 2013, Lake Tahoe, Nevada, United States},
  pages 3111--3119.

\bibitem[{Moon et~al.(1995)Moon, Rajagopalan, and
  Lall}]{moon1995_mi_estimation_kernel}
Young-Il Moon, Balaji Rajagopalan, and Upmanu Lall. 1995.
\newblock Estimation of mutual information using kernel density estimators.
\newblock \emph{Physical Review E}, 52(3):2318.

\bibitem[{Paninski(2003)}]{paninski_mi_estimation_histogram}
Liam Paninski. 2003.
\newblock \href {https://doi.org/10.1162/089976603321780272} {Estimation of
  entropy and mutual information}.
\newblock \emph{Neural Comput.}, 15(6):1191--1253.

\bibitem[{Peng et~al.(2005)Peng, Long, and Ding}]{peng_mi_estimation}
Hanchuan Peng, Fuhui Long, and Chris H.~Q. Ding. 2005.
\newblock \href {https://doi.org/10.1109/TPAMI.2005.159} {Feature selection
  based on mutual information: Criteria of max-dependency, max-relevance, and
  min-redundancy}.
\newblock \emph{{IEEE} Trans. Pattern Anal. Mach. Intell.}, 27(8):1226--1238.

\bibitem[{Pennington et~al.(2014)Pennington, Socher, and
  Manning}]{jeffrey_glove}
Jeffrey Pennington, Richard Socher, and Christopher~D. Manning. 2014.
\newblock \href {https://doi.org/10.3115/v1/d14-1162} {Glove: Global vectors
  for word representation}.
\newblock In \emph{Proceedings of the 2014 Conference on Empirical Methods in
  Natural Language Processing, {EMNLP} 2014, October 25-29, 2014, Doha, Qatar,
  {A} meeting of SIGDAT, a Special Interest Group of the {ACL}}, pages
  1532--1543. {ACL}.

\bibitem[{Perozzi et~al.(2014)Perozzi, Al{-}Rfou, and Skiena}]{deepwalk}
Bryan Perozzi, Rami Al{-}Rfou, and Steven Skiena. 2014.
\newblock \href {https://doi.org/10.1145/2623330.2623732} {Deepwalk: online
  learning of social representations}.
\newblock In \emph{The 20th {ACM} {SIGKDD} International Conference on
  Knowledge Discovery and Data Mining, {KDD} '14, New York, NY, {USA} - August
  24 - 27, 2014}, pages 701--710. {ACM}.

\bibitem[{Pimentel et~al.(2020{\natexlab{a}})Pimentel, Saphra, Williams, and
  Cotterell}]{pimentel-etal-2020-pareto}
Tiago Pimentel, Naomi Saphra, Adina Williams, and Ryan Cotterell.
  2020{\natexlab{a}}.
\newblock \href {https://doi.org/10.18653/v1/2020.emnlp-main.254} {Pareto
  probing: Trading off accuracy for complexity}.
\newblock In \emph{Proceedings of the 2020 Conference on Empirical Methods in
  Natural Language Processing, {EMNLP} 2020, Online, November 16-20, 2020},
  pages 3138--3153. Association for Computational Linguistics.

\bibitem[{Pimentel et~al.(2020{\natexlab{b}})Pimentel, Valvoda, Maudslay,
  Zmigrod, Williams, and Cotterell}]{mi_probe}
Tiago Pimentel, Josef Valvoda, Rowan~Hall Maudslay, Ran Zmigrod, Adina
  Williams, and Ryan Cotterell. 2020{\natexlab{b}}.
\newblock \href {https://doi.org/10.18653/v1/2020.acl-main.420}
  {Information-theoretic probing for linguistic structure}.
\newblock In \emph{Proceedings of the 58th Annual Meeting of the Association
  for Computational Linguistics, {ACL} 2020, Online, July 5-10, 2020}, pages
  4609--4622. Association for Computational Linguistics.

\bibitem[{Pourdamghani et~al.(2014)Pourdamghani, Gao, Hermjakob, and
  Knight}]{nima_amr_aligner}
Nima Pourdamghani, Yang Gao, Ulf Hermjakob, and Kevin Knight. 2014.
\newblock \href {https://doi.org/10.3115/v1/d14-1048} {Aligning english strings
  with abstract meaning representation graphs}.
\newblock In \emph{Proceedings of the 2014 Conference on Empirical Methods in
  Natural Language Processing, {EMNLP} 2014, October 25-29, 2014, Doha, Qatar,
  {A} meeting of SIGDAT, a Special Interest Group of the {ACL}}, pages
  425--429. {ACL}.

\bibitem[{Recht and R{\'e}(2012)}]{random_sample_training_recht2012beneath}
Benjamin Recht and Christopher R{\'e}. 2012.
\newblock Beneath the valley of the noncommutative arithmetic-geometric mean
  inequality: conjectures, case-studies, and consequences.
\newblock \emph{arXiv preprint arXiv:1202.4184}.

\bibitem[{Reid et~al.(2020)Reid, Marrese{-}Taylor, and Matsuo}]{elmo}
Machel Reid, Edison Marrese{-}Taylor, and Yutaka Matsuo. 2020.
\newblock \href {https://doi.org/10.18653/v1/2020.emnlp-main.513} {{VCDM:}
  leveraging variational bi-encoding and deep contextualized word
  representations for improved definition modeling}.
\newblock In \emph{Proceedings of the 2020 Conference on Empirical Methods in
  Natural Language Processing, {EMNLP} 2020, Online, November 16-20, 2020},
  pages 6331--6344. Association for Computational Linguistics.

\bibitem[{Reif et~al.(2019)Reif, Yuan, Wattenberg, Vi{\'{e}}gas, Coenen,
  Pearce, and Kim}]{emily_geometric_bert}
Emily Reif, Ann Yuan, Martin Wattenberg, Fernanda~B. Vi{\'{e}}gas, Andy Coenen,
  Adam Pearce, and Been Kim. 2019.
\newblock \href
  {https://proceedings.neurips.cc/paper/2019/hash/159c1ffe5b61b41b3c4d8f4c2150f6c4-Abstract.html}
  {Visualizing and measuring the geometry of {BERT}}.
\newblock In \emph{Advances in Neural Information Processing Systems 32: Annual
  Conference on Neural Information Processing Systems 2019, NeurIPS 2019,
  December 8-14, 2019, Vancouver, BC, Canada}, pages 8592--8600.

\bibitem[{Rogers et~al.(2020)Rogers, Kovaleva, and
  Rumshisky}]{primer_bert_RogersKR20}
Anna Rogers, Olga Kovaleva, and Anna Rumshisky. 2020.
\newblock \href {https://transacl.org/ojs/index.php/tacl/article/view/2257} {A
  primer in bertology: What we know about how {BERT} works}.
\newblock \emph{Trans. Assoc. Comput. Linguistics}, 8:842--866.

\bibitem[{Ross(2014)}]{mi_estimation_discrete_continuous}
Brian~C Ross. 2014.
\newblock Mutual information between discrete and continuous data sets.
\newblock \emph{PloS one}, 9(2):e87357.

\bibitem[{Steuer et~al.(2002)Steuer, Kurths, Daub, Weise, and
  Selbig}]{steuer_mi_estimation_histogram}
Ralph~E. Steuer, J{\"{u}}rgen Kurths, Carsten~O. Daub, Janko Weise, and Joachim
  Selbig. 2002.
\newblock The mutual information: Detecting and evaluating dependencies between
  variables.
\newblock In \emph{Proceedings of the European Conference on Computational
  Biology {(ECCB} 2002), October 6-9, 2002, Saarbr{\"{u}}cken, Germany}, pages
  231--240.

\bibitem[{Tang et~al.(2015)Tang, Qu, Wang, Zhang, Yan, and
  Mei}]{line_TangQWZYM15}
Jian Tang, Meng Qu, Mingzhe Wang, Ming Zhang, Jun Yan, and Qiaozhu Mei. 2015.
\newblock \href {https://doi.org/10.1145/2736277.2741093} {{LINE:} large-scale
  information network embedding}.
\newblock In \emph{Proceedings of the 24th International Conference on World
  Wide Web, {WWW} 2015, Florence, Italy, May 18-22, 2015}, pages 1067--1077.
  {ACM}.

\bibitem[{Tenney et~al.(2019{\natexlab{a}})Tenney, Das, and
  Pavlick}]{tenney-etal-2019-bert}
Ian Tenney, Dipanjan Das, and Ellie Pavlick. 2019{\natexlab{a}}.
\newblock \href {https://doi.org/10.18653/v1/p19-1452} {{BERT} rediscovers the
  classical {NLP} pipeline}.
\newblock In \emph{Proceedings of the 57th Conference of the Association for
  Computational Linguistics, {ACL} 2019, Florence, Italy, July 28- August 2,
  2019, Volume 1: Long Papers}, pages 4593--4601. Association for Computational
  Linguistics.

\bibitem[{Tenney et~al.(2019{\natexlab{b}})Tenney, Xia, Chen, Wang, Poliak,
  McCoy, Kim, Durme, Bowman, Das, and Pavlick}]{edge_probe}
Ian Tenney, Patrick Xia, Berlin Chen, Alex Wang, Adam Poliak, R.~Thomas McCoy,
  Najoung Kim, Benjamin~Van Durme, Samuel~R. Bowman, Dipanjan Das, and Ellie
  Pavlick. 2019{\natexlab{b}}.
\newblock \href {https://openreview.net/forum?id=SJzSgnRcKX} {What do you learn
  from context? probing for sentence structure in contextualized word
  representations}.
\newblock In \emph{7th International Conference on Learning Representations,
  {ICLR} 2019, New Orleans, LA, USA, May 6-9, 2019}. OpenReview.net.

\bibitem[{Voita and Titov(2020)}]{mdl_probe}
Elena Voita and Ivan Titov. 2020.
\newblock \href {https://doi.org/10.18653/v1/2020.emnlp-main.14}
  {Information-theoretic probing with minimum description length}.
\newblock In \emph{Proceedings of the 2020 Conference on Empirical Methods in
  Natural Language Processing, {EMNLP} 2020, Online, November 16-20, 2020},
  pages 183--196. Association for Computational Linguistics.

\bibitem[{Wu et~al.(2021)Wu, Peng, and Smith}]{wu2020infusing}
Zhaofeng Wu, Hao Peng, and Noah~A. Smith. 2021.
\newblock \href {https://transacl.org/ojs/index.php/tacl/article/view/2603}
  {Infusing finetuning with semantic dependencies}.
\newblock \emph{Trans. Assoc. Comput. Linguistics}, 9:226--242.

\end{thebibliography}

\appendix
\clearpage

\section{Details of Graph Embedding} \label{appendix:ge}
In this section, we present the technique details of the graph embedding approach, as well as parameters. Given a graph such as syntax tree and semantic graph, we first run random walk algorithm on it to sample walk paths. The random walk strategy is simple, each time a neighbor of current node is selected from its neighbor set based on uniform distribution. For each node, the length of random walk path is $10$. And the repeat time is $100$. In general, each node has $100$ different walk paths with length $10$. Then we put those paths into Word2vec model~\citep{word2vec}, with window size equal to $2$, since we only want graph embeddings to capture the one-hop neighborhood relationships. The hidden states, in other words graph embeddings are vectors with $128$ dimensions.

\section{Details of MI Estimation} \label{appendix:mi}
In this section we present technique details about MI estimation, such as the neural network model design and parameters. There are two terms in the objective function~\ref{eq:obj}: one is about joint distribution $$\mathbb{E}_{\mathbb{P}^{(n)}_{\mathcal{X}\mathcal{Z}}}[T_{\theta}]$$ and another is about marginal distribution $$\log(\mathbb{E}_{\mathbb{P}^{(n)}_{\mathcal{X}} \otimes \mathbb{P}^{(n)}_{\mathcal{Z}}}[e^{T_{\theta}}]).$$ For the joint distribution part, we concatenate the graph embeddings $\mathcal{Z}$ and word representations $\mathcal{X}$ first, and then put them into our designed neural network to compute a scalar. Then the average value of the scalar is computed. For the marginal part, we randomly shuffle the representations $\mathcal{X}$. After random shuffle, there is no dependency between $\mathcal{X}$ and $\mathcal{Z}$ anymore. Then, we put the concatenation of the shuffled representations and graph embeddings into the neural network to get another scalar. We take the exponential value of that scalar and take the average value for the whole dataset.

As aforementioned, the selection of model size is a tradeoff. In our experiments, we design an MLP model with two layers for MI estimation. The first layer is linear without nonlinear activation function, to encode graph embeddings and word representations into same space, with $64$ dimensions. Then we concatenate those two hidden states and put them through a nonlinear layer to get a scalar. For example, we have one sentence with $10$ words. Assume we can get graph embeddings with size $10* 128$, and word representations of BERT with size $10* 768$. Then we use a linear function to map those two vectors into hidden space, say with $32$ dimensions. Then we concatenate those two embeddings as a $10* 64$ matrix, and use one extra linear function with nonlinear activation functions to map it as $10* 1$ matrix. Then we can get the compression lemma lower bound as the mean value of the $10* 1$ matrix, which is the mutual information estimation that we want. 

The loss is defined directly as the minus value of objective function. With stochastic gradient decent, we can maximize the lower bound to get the estimation. About the mini-batch, since the document contains many sentences, we select one sentence as one min-batch to optimize the neural network. 

The reason why we treat one sentence as one minibatch is that we get word representations of BERT and graph embeddings in that way. One sentence has a complete syntax tree structure, and getting word representations with one sentence in BERT can make attention computed within the sentence. Another reason is that using two sentences as input may exceed the maximum BERT input size: $512$ tokens sometimes. However, if we use mini-batch to estimate the mutual information, it brings errors. The reason is that if we want to estimate the mutual information between $\mathcal{X}$ and $\mathcal{Z}$, the expectation should be all the data points that we know. But here we use minibatch to calculate the expectation for one batch only. To alleviate this error, as introduced in~\citet{mine}, we select small learning rate to keep the error small.

\section{Details of MLP Models} \label{appendix:mlp}
This section introduces how to use MLP to do link prediction task, as well as the details about MLPs. For one sentence, given its graph embeddings, we simply use MLP to calculate a score for all node pairs, and then compare with the ground-truth graph with the predicted distribution vector. AUC score is computed based on the distribution vector and ground-truth vector. Note that since the graph is very sparse, it makes the task very difficult. Generally, the task can be regarded as a binary classification task with an extremely unbalanced data distribution. 

For the details, linear MLP simply predicts the graph by the concatenation of two input vectors to decide whether there is an edge between them. For MLPs with hidden layers, The concatenated vectors are through non-linear layers first, then the final output layer is linear. The dimension of all hidden states is $128$. And the learning rate is $10^{-4}$.

\section{Reliability of MI Estimation} \label{appendix:mi_estimation}
MI estimation for features in high-dimensional space is difficult. To prove that our estimated MI values are quite accurate, we test the MI estimation method~\citep{mine} on sets of graph embeddings with different levels of dependencies.
To have that, we add noise on graph embeddings as $\mathcal{Z}'$. $\mathcal{Z}$ and $\mathcal{Z}'$ can have different dependencies based on the added noise rate. Noise vectors are sampled from a standard Gaussian independently. We test the estimation for various levels of noise, from original graph embeddings $\mathcal{Z}$ to the condition that $100\%$ signals are noise $\sigma$. For example, $40\%$ means that for each graph embedding $Z' = 60 \% \times Z + 40\% \times \sigma$. Then, we calculate $\hat{I}(\mathcal{Z}';\mathcal{Z})$ to see whether the values is small with large noise added.

\begin{table}[ht]
	\caption{MI value with different level of noise}
	\label{tab:mi_test}
	\vspace{-2mm}
	\scalebox{0.69}{
		\begin{tabular}{c|c|c}
			\hline
			\toprule
			\diagbox{Noise ratio (\%)} {MI percentage (\%)}{Structure} & Syntactic tree & Semantic graph \\
			\midrule
			\hline
			 0 & 100.00 & 100.00 \\
			 10 & 91.84 & 80.40 \\
			 20 & 74.10 & 63.29 \\
			 30 & 62.23 & 51.88 \\
			 40 & 49.53 & 32.70 \\
			 50 & 38.61 & 20.44 \\
			 60 & 29.46 & 9.36 \\
			 70 & 22.09 & 3.74 \\
			 80 & 12.25 & 0.94 \\
			 90 & 2.07 & 0.02 \\
			 100 & -0.03 & -0.06 \\
			\bottomrule
			\hline
		\end{tabular}}
\end{table}
Table~\ref{tab:mi_test} presents the exact values. To make it more readable, we report the MI percentage of $\hat{I}(\mathcal{Z}';\mathcal{Z})/\hat{I}(\mathcal{Z};\mathcal{Z})$. From the results, we find that for the two structures, results are not very similar. But the general tendencies are consistent, where less dependencies caused by larger noise have smaller MI values.
Note that when the noise rate is $100\%$, $\hat{I}(\mathcal{Z}',\mathcal{Z})$ degenerates to the lower bound $\hat{I}(\mathcal{R},\mathcal{Z})$. As mentioned before, the absolute value of it represents the gap between the estimated MI and ground-truth MI, which is very small (less than $10^{-3} \times \hat{I}(\mathcal{Z};\mathcal{Z})$). It also proves that our MI estimations are reliable.

\section{Statistics of Penn Treebank and AMR Bank} \label{appendix:ptb}
We provide the relation number and connected word number of Penn Treebank dataset and AMR Bank in this section. The word number for POS tags of Penn Treebank is also provided. The statistics are for the whole dataset.
\begin{table}[htbp]
	\caption{Penn Treebank statistics of syntax relations}
	\vspace{-2mm}
	\scalebox{0.58}{
		\begin{tabular}{c|c|c|c}
			\hline
			\toprule
			\diagbox{Relation} {Statistic} & $\#$ of Relations & $\%$ of Relations & $\#$ of Connected Words \\
			\midrule
			\hline
            punct &	121,395 &	11.60 &	193,648 \\
            prep &	100,997 &	9.648 &	189,783 \\
            pobj &	98,586 &	9.418 &	197,164 \\
            det &	86,228 &	8.237 &	172,446 \\
            nn &	81,381 &	7.774 &	143,248 \\
            nsubj &	73,802 &	7.050 &	147,498 \\
            amod &	66,381 &	6.341 &	125,119 \\
            root &	43,948 &	4.198 &	87,896 \\
            dobj &	43,054 &	4.113 &	85,997 \\
            aux &	37,267 &	3.560 &	73,436 \\
            $\#$ others & 26,1791 & 25.01 & 509,641 \\

			\bottomrule
			\hline
		\end{tabular}}
\end{table}
We only report details of relations ranked top $10$. However, there are other $35$ types of relations, which are categorized together in type $\#$ others.
\begin{table}[htbp]
	\caption{Penn Treebank statistics of POS tags}
	\vspace{-2mm}
	\scalebox{0.955}{
		\begin{tabular}{c|c|c}
			\hline
			\toprule
			\diagbox{Tags} {Statistic} & $\#$ of Words & $\%$ of Words \\
			\midrule
			\hline
            NN & 146,228 & 15.89 \\
            IN & 108,434 & 11.78 \\
            NNP & 101,427 & 11.02 \\
            DT & 90,158 & 9.798 \\ 
            JJ & 67,396 & 7.324 \\
            NNS & 65,867 & 7.158 \\
            CD & 40,337 & 4.384 \\
            RB & 34,331 & 3.731 \\
            VBD & 33,430 & 3.633 \\
            VB & 29,001 & 3.152 \\
            $\#$ others & 203,578 & 22.12 \\
			\bottomrule
			\hline
		\end{tabular}}
\end{table}
For the POS tagging statistics, we also only present tags with word number ranked top $10$. There are still $28$ types of POS tags that are catergorized into one type $\#$ others.
\begin{table}[htbp]
	\caption{AMR bank statistics of relations}
	\vspace{-2mm}
	\scalebox{0.86}{
		\begin{tabular}{c|c|c}
			\hline
			\toprule
			\diagbox{Relation} {Statistic} & $\#$ of Relations & $\%$ of Relations \\
			\midrule
			\hline
            arg & 409,322 & 58.11  \\
            general & 208,287 & 29.57  \\
            op & 67,307 & 9.556  \\
            quantities & 13,092 & 1.859  \\
            others & 5,216 & 0.7406 \\
            date & 1,114 & 0.1582 \\
			\bottomrule
			\hline
		\end{tabular}}
\end{table}

The AMR graphs are different from syntax trees. The relations of AMR graphs can be classified into $6$ groups, and each group contains many types. Specifically, the group \textit{general} includes: ``accompanier'', ``age'', ``beneficiary'', ``concession'', ``condition'', ``consist'', ``degree'', ``destination'', ``direction'', ``domain'', ``duration'', ``example'', ``extent'', ``frequency'', ``instrument'', ``location'', ``manner'', ``medium'', ``mod'', ``name'', ``part'', ``path'', ``polarity'', ``poss'', ``purpose'', ``source'', ``subevent'', ``subset'', ``time'', ``topic'', ``value'', ``ord'', and ``range''. And the \textit{quantities} group includes ``quant'', ``scale'', and ``unit''.

For the description, \textit{arg} represents frame arguments, following PropBank conventions. \textit{general} are composed of a set of general semantic relations. \textit{op} means the relations for lists. Similarly, \textit{quantities} are relations for quantities. And \textit{date} contains relations for date-entities.

\section{Probe Specific Semantic Relations} \label{appendix:semantic_local}
\begin{figure}[ht]
	\centering
	\includegraphics[scale=0.25]{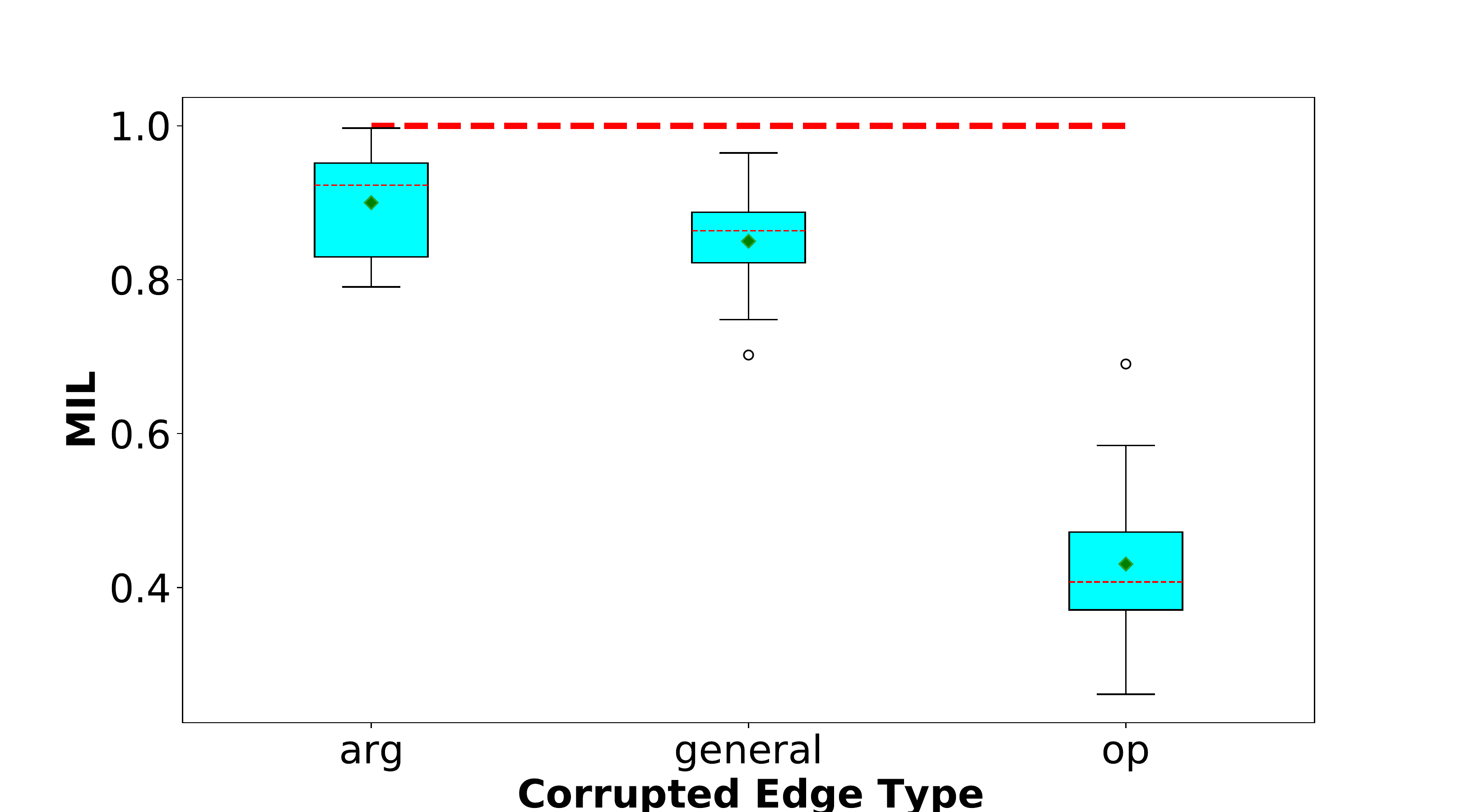}
	\vspace{-6.mm}
	\caption{$MIL$ scores of probing $5$ types of localized semantic structure for word representations in BERT-base (output layer). The local structure is decided by the labels attached on edges in AMR graph.}
	\label{fig:noisy_amr}
\end{figure}
For semantic structure, we run the perturbation experiment in a similar way. Different from syntax trees, the relations of AMR graphs can be grouped into $6$ types. And number distribution is not very even. Thus, we corrupt the graph embeddings with $50\%$ noise. Other settings are similar to that of syntactic structures. From the results, we can found that BERT encodes more structure information about \textit{arg} and \textit{general} relations. While for the \textit{op} relations, which represents the relations for lists, are not well encoded.

\section{Probing Localized Information with Accuracy} \label{appendix:perturbation_accuracy}
\begin{figure*}[htbp]
\centering
\subfigure{
\begin{minipage}[t]{0.485\linewidth}
	\includegraphics[scale=0.595]{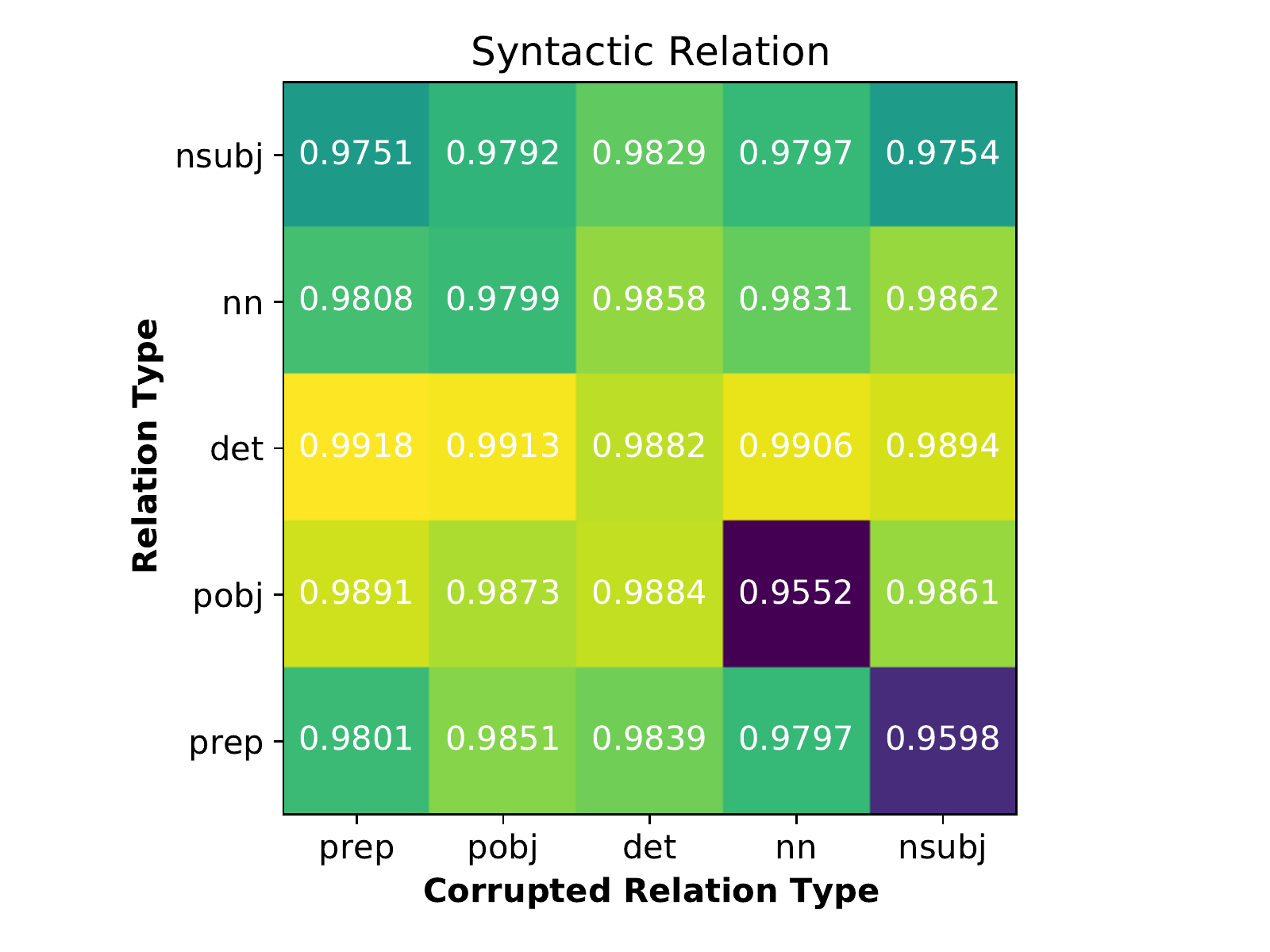}
	\label{fig:random_ptb}
\end{minipage}
}
\subfigure{
\begin{minipage}[t]{0.485\linewidth}
	\includegraphics[scale=0.595]{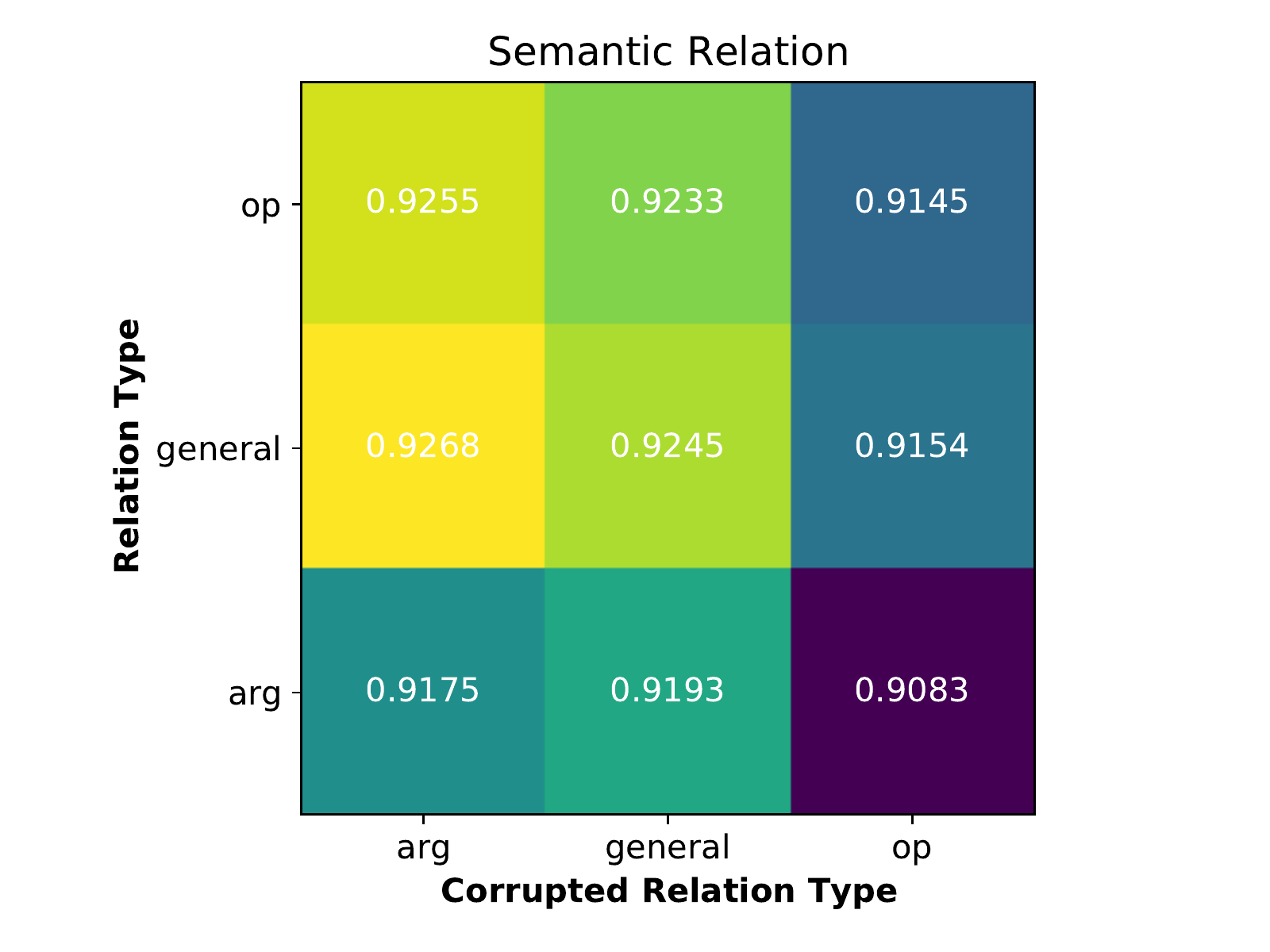}
	\label{fig:random_amr}
\end{minipage}
}
\centering
\vspace{-6.mm}
\caption{Probe specific syntactic and semantic relations}
\label{fig:random}
\end{figure*}
Similar to our localized probing experiments, we add perturbations on word representations. Specifically, we corrupt word representations equally and with same number for each relation type. Then we train an MLP with $5$ hidden layers to predict entire structures with the corrupted word representations. AUC scores of all relation types are calculated. As in the \we, we only report $5$ types of relations for syntactic and $3$ types for semantic structures. Results are shown in Figure~\ref{fig:random_ptb} and Figure~\ref{fig:random_amr}. We can find that the probe accuracies 
are very unusual. First, corrupt one type of relations, the accuracies for other types of relations change significantly. Besides, the MLP is trained with corrupted relations e.g., \textit{nsubj} while predicts \textit{prep} with worst AUC score. The results also prove the point that prediction decisions are not based purely on the structure but rather on spurious heuristics~\citep{lovering2021informationtheoretic}.

\end{document}